\documentclass[a4paper,fleqn]{cas-dc}

\usepackage[numbers]{natbib}
\usepackage{xcolor}

\begin{document}
\let\WriteBookmarks\relax
\def\floatpagepagefraction{1}
\def\textpagefraction{.001}

\shorttitle{On-device edge learning for IoT data streams: a survey}

\shortauthors{Louren\c{c}o et~al.}

\title [mode = title]{On-device edge learning for IoT data streams: a survey}


\tnotetext[1]{This work was funded by the EU, through the Portuguese Republic’s Recovery and Resilience Plan, within the project PRODUTECH R3. It was also funded by the Portuguese Foundation for Science and Technology under project doi.org/10.54499/UIDP/00760/2020 and Ph.D. scholarship PRT/BD/154713/2023.}

\author[1]{Afonso Louren\c{c}o}[orcid=0000-0002-3465-3419]

\cormark[1]


\ead{fonso@isep.ipp.pt}

%\ead[url]{www.cvr.cc, cvr@sayahna.org}

%\credit{Conceptualization of this study, Methodology, Software}

\affiliation[1]{organization={GECAD, ISEP, Polytechnic of Porto},
    addressline={Rua Dr. António Bernardino de Almeida}, 
    city={Porto},
    postcode={4249-015}, 
    country={Portugal}}

\author[1]{João Rodrigo}[orcid=0009-0008-8702-3876]

\author[2]{João Gama}[orcid=0000-0003-3357-1195]
   
%\ead{cvr3@sayahna.org}
%\ead[URL]{www.sayahna.org}

%\credit{Data curation, Writing - Original draft preparation}

\affiliation[2]{organization={INESC TEC, FEP, University of Porto},
    addressline={Rua Dr. Roberto Frias}, 
    city={Porto},
    postcode={4200-465}, 
    country={Portugal}}

\author[1]{Goreti Marreiros}[orcid=0000-0003-4417-8401]
%\cormark[2]
%\fnmark[1,3]
%\ead{rishi@stmdocs.in}
%\ead[URL]{www.stmdocs.in}

\cortext[cor1]{Corresponding author}

%\nonumnote{This note has no numbers. In this work we demonstrate $a_b$ the formation Y\_1 of a new type of polariton on the interface between a cuprous oxide slab and a polystyrene micro-sphere placed on the slab.}

\begin{abstract}
This literature review explores continual learning methods for on-device training in the context of neural networks (NNs) and decision trees (DTs) for classification tasks on smart environments. We highlight key constraints, such as data architecture (batch vs. stream) and network capacity (cloud vs. edge), which impact TinyML algorithm design, due to the uncontrolled natural arrival of data streams. The survey details the challenges of deploying deep learners on resource-constrained edge devices, including catastrophic forgetting, data inefficiency, and the difficulty of handling IoT tabular data in open-world settings. While decision trees are more memory-efficient for on-device training, they are limited in expressiveness, requiring dynamic adaptations, like pruning and meta-learning, to handle complex patterns and concept drifts. We emphasize the importance of multi-criteria performance evaluation tailored to edge applications, which assess both output-based and internal representation metrics. The key challenge lies in integrating these building blocks into autonomous online systems, taking into account stability-plasticity trade-offs, forward-backward transfer, and model convergence.
\end{abstract}

% \includegraphics{figs/grabs.pdf}

\iffalse
 \begin{highlights}
\item On-device training for classification with neural networks and decision trees
\item Constraints of data architecture (batch/stream) and network capacity (cloud/edge)
\item NN compression, catastrophic forgetting, open-world setting, and IoT tabular data
\item DT adaptations, focusing on memory, unsupervised ensembles, and structural changes
\item Multi-objective evaluation metrics for continual learning at edge devices
\end{highlights}
\fi

% Keywords

\maketitle

\section{Introduction}

In today's interconnected world, nearly every electronic device is transmitting data over the internet, whether intentionally or not. The Internet of Things (IoT) continues to evolve, enabling the optimization of processes across a wide range of domains \cite{warden2019tinyml}. While initially, only servers had the necessary computing power for advanced analytics, as technology evolved, smaller devices had competing power for some applications, eliminating network delays in areas where critical decisions must be made in an instant. This shift in data generation and utilization gives rise to two key paradigms: ubiquitous computing, which refers to the pervasive presence of processing power throughout our environments, making them more interconnected and intelligent; and edge computing, which emphasizes the location of data processing by moving computation closer to the data source, reducing reliance on centralized cloud infrastructures. In particular, due to the widespread adoption of relational databases in these domains, tabular data is the dominant modality in these IoT applications. Organized into rows and columns, consisting of distinct features that are typically continuous, categorical, or ordinal, data arrives continuously as an infinite data stream.

\textbf{Lifelong learning.}  This challenges the traditional batch learning paradigm, which assumes the availability of all training data upfront and its independent and identically distributed nature. Since a model only has access to the current data in an individual phase of the learning cycle, it is prone to overfit on the currently available data, with catastrophic forgetting of previous data \cite{schwarz2018progress}. Conversely, holding on to outdated knowledge can hinder the model ability to adapt fast and effectively learn from new data. Naturally, this stability-plasticity dilemma, is related to the challenge of learning invariant representations where the model exhibits both forward and backward transfer, in which learning a new concept not only takes advantage but also benefits from the knowledge extracted from old concepts.

\textbf{Resource efficiency.} Simultaneously, another challenge comes from the practical memory constraint of handling very long non-stationary sequences, without storing data. Under such a constraint, one must develop effective surrogate learning objectives that can account for past errors with high resource efficiency, quantified through computational overhead or energy consumption, and as a function of parameter growth, storage of data samples, or a pool of model copies. In particular, the TinyML community defines such requirements around a memory of 100KB to 1MB, and a processing power between 10MHz and 100MHz \cite{warden2019tinyml}. Moreover, to handle high-speed data streams, the model should work in a limited amount of time, ensuring it can incorporate new information as it arrives, predict, and update the model at any point. 

\textbf{Two communities.} To tackle these challenges of bringing advanced AI to edge environments, two parallel research communities have emerged, with their own predefined data schemas and prediction tasks specifically tailored for on-device edge training. Both these communities recognize one cannot repeatedly access data from streams and reconstruct new models from scratch. Instead, they propose some form of stateful learning, detecting patterns without the need to store all the data and pass through them multiple times.

\textbf{Neural networks.} The first community focuses on deep learning algorithms, being more fragmented in its solutions for edge requirements, mostly identified by “online continual learning”, "TinyML", "lightweight AI", or “task-free learning” \cite{warden2019tinyml,verwimp2023continual}. It mostly focuses on including mechanisms to shrink the size of NNs, such as pruning, and quantization, \cite{howard2017mobilenets, iandola2016squeezenet}, and incrementally processing mini-batches of data from varying distributions, precluding multiple epochs of offline training and the storage of data \cite{kirkpatrick2017overcoming,caccia2022new}. The major challenge of these methods are catastrophic forgetting, and the very slow convergence rate in performing stochastic updates to these entangled model structures. Since no weights are fixed, previously learned knowledge can easily be overwritten when training on new data from a different distribution. Moreover, this plasticity does not necessarily imply that NNs can always learn new data efficiently. In fact, they become extremely data inefficient as more data is learned, requiring various passes of old data to mitigate the interference between old and new representations. Moreover, as deep learning architectures have been crafted with inductive biases for matching invariances and homogeneous data, they struggle to learn the irregular patterns of IoT tabular data \cite{grinsztajn2022tree}.

\textbf{Decision trees.} The second community recognizes itself by keywords such as “data stream mining”, “concept drift”, and “online learning” \cite{domingos2000mining,lourenco2025dfdt}. It mostly focuses on ensembles of resource-efficient tree-based learners that rely on statistical bounds of incrementally computed information gain measures to determine whether the observed utility of a split is statistically significant \cite{gomes2017adaptive,gunasekara2024gradient}. The major challenge of these methods is a lack of plasticity. As the tree grows from the root node, the descendant nodes subsequently get fixed to covering particular sub-spaces of the parent node. Consequently, this community focuses on developing advanced methods to not only carefully select when to expand, but also to prune affected parent nodes, instead of a complete replacement of the base learner \cite{domingos2001catching}. Overall, these shallow learners possess a faster online convergence rate, thanks to their simple model structures with a small number of trainable parameters. However, due to their limited learning capacity, they can end up with inferior performance than deep learners, whose low-dimensional hidden representations allow better feature interplay on complex raw inputs.

\textbf{This survey.} Despite this extensive research from both communities, a key gap exists in integrating these areas, particularly with respect to memory optimization and the handling of tabular data streams for on-device training. While continuous learning algorithms have made strides in addressing the challenges of catastrophic forgetting and learning from sequential data, they often do so without considering the unique constraints of memory-limited edge devices that process IoT data \cite{verwimp2023continual}. Conversely, efforts in reducing computational overhead through pruning or quantization, are often disconnected from the challenges of incrementally incorporating and preserving knowledge from heterogeneous, feature-rich, and dynamic nature of IoT data. To bridge these areas, this survey proposes to answer the following research questions, each one being respectively addressed by a section:

\begin{enumerate}
    \item[\textbf{RQ1:}] What are the fundamental requirements for developing on-device training solutions? How do different data architectures (batch vs stream) and network capacities (cloud vs edge) influence algorithm design?
    
    \item[\textbf{RQ2:}] How can NNs be deployed at the edge while continuously learning, under resource constraints, open-world settings, and IoT tabular data?
    
    \item[\textbf{RQ3:}] How to make DTs viable for on-device training, in terms of dynamic architectures, memory-efficient growth, ensemble learning, and without supervision?
    
    \item[\textbf{RQ4:}] How can the suitability of an on-device training strategy be assessed, and what evaluation metrics ensure robust model performance before deployment?
\end{enumerate}

\section{On-device edge learning}
\label{sec:edge}

\begin{figure}[t!]
    \centering
    \includegraphics[width=\linewidth]{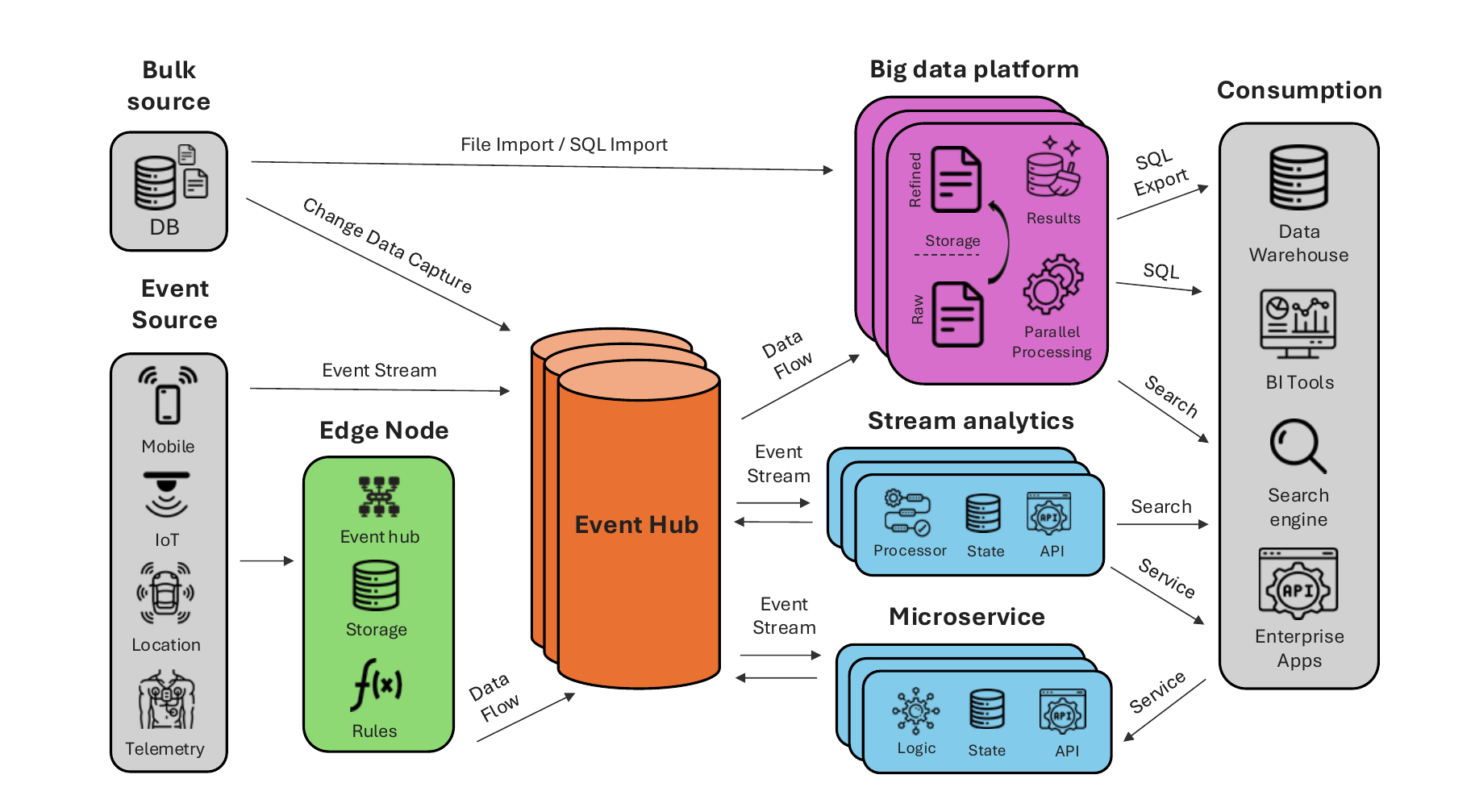}
    \caption{Data mining architecture}
    \label{fig:data-processing}
\end{figure}

To address RQ1, this section aims at redefining the requirements for developing on-device training solutions. For this purpose, two major drivers of edge applications were discussed, along with consequence AI algorithm design implications. On one hand, how contrasting approaches to processing data, namely in batch or stream architectures, lead to different AI training paradigms, with an inference engine either trained offline or online. On the other hand, how different network capacities, caused by moving data processing closer or further from where the data is produced, lead to different AI resources paradigms in CloudML, MobileML and TinyML. These topics will refer to Figure \ref{fig:data-processing}.

\subsection{Data architecture}

In the IoT environment, several data processing architectures are employed, each designed to address the specific characteristics of the data. These include batch processing for static datasets, which is commonly implemented using Extract, Transform, and Load (ETL) pipelines and data lakes; batch processing for stream data, often facilitated through event hubs and publish-subscribe (pub-sub) communication; and stream processing architectures that can handle both batch and stream data, typically leveraging change data capture (CDC) techniques.

\textbf{\textcolor{magenta}{Big data platform.}} Batch architectures are most commonly use in applications that do not require immediate responses, but instead focus on algorithm performance and understanding patterns within the data \cite{warren2015big}.  In a typical batch processing pipeline, data is handled through ETL operations \cite{van2020evaluation}. Once transformed, the data is saved in transactional databases, such as MySQL, where it is structured into tables and then into schemas optimized for complex analytics queries \cite{debnath2008sard}. To deal with the vast amounts of diverse, unstructured information, raw data can be alternatively stored in data lakes \cite{warren2015big}. Typically, a batch system retrieves records on a scheduled basis, e.g. monthly, via APIs. Once the data request is fulfilled, all files are imported, the connection is halted, and the processing begins. Technologies such as MapReduce or Spark are typically used in the processing layer, after which results are stored and delivered to the final consumer \cite{warren2015big}. However, while data is processed in bulk its collection should take into account the sequential nature of data, such as in engine readings. To address this issue, a data streaming processor can act as a mediator, allowing downstream systems to consume events at their own pace and in the correct order \cite{lin2017lambda}. Unlike data lakes, where all services read from the same queue, thus creating complications due to the need for additional filtering, a data streaming processor organizes messages into queues, storing them as if they were in a long log file, with data being directed into separate streams or topics \cite{kreps2014questioning}. This approach models data as transient streams rather than persistent tables, enabling services to subscribe only to their specific topics of interest \cite{warren2015big}. 

\textbf{\textcolor{orange}{Event hub.}} For this purpose, a streaming processor uses the pub-sub communication method. In contrast to traditional web communication via APIs, where the system synchronously sends a request and waits for a response, pub-sub enables asynchronous communication between multiple systems, allowing large volumes of data to be generated simultaneously. This effectively decouples data sources from data consumers. Data is divided into topics or profiles, and consumers subscribe to these topics. When a new data record or event is generated, it is published within the topic, allowing subscribers to consume the data at their own pace. This method enables systems to handle thousands of events per second without the delays associated with synchronous responses \cite{lin2017lambda}. However, data is still processed in batches at regular intervals. Thus, to alleviate the consequent batch processing bottleneck, data is buffered in an event hub. Ultimately, the issue of high latency remains. By the time files are obtained from the source, they are already outdated. Loading them into systems such as Hadoop, and subsequent processing steps, such as using MapReduce or Spark, further delay the data's freshness \cite{kreps2014questioning, van2020evaluation}. As a result, the final output can not reflect up-to-date results, with IoT data losing value over time. While event hubs provide an effective buffering solution, the asynchronous nature of data storage and analytics necessitates innovative approaches to ensure timely data processing and retrieval \cite{van2020evaluation}.

\textbf{\textcolor{cyan}{Stream analytics and microservices.}} To address this, streaming architectures emerged, where analytics are performed on each event as it arrives, e.g. when detecting an anomaly, send an event back to the event stream for further processing or action by another service. For this purpose, streaming architectures typically consist of multiple services, each handling specific tasks, rather than a single monolithic system. This separation of concerns is crucial for scalability and maintainability. Microservices can consume data from event tables and produce new events back to the event stream, e.g. a dispatch service, using a ML model to analyze vehicle speed and location in real-time, can continuously send adjusted results on estimated arrival times. In such systems, dealing with limited event information to minimize message size is common. However, additional information might be required to enrich stream analytics \cite{ankorion2005change}. For instance, in a scenario where vehicles are monitored in real time, the system might transmit data such as the vehicle ID, driver ID, vehicle position, and some driver behavior metrics. While this event data provides valuable real-time insights, more in-depth analytics may require supplementary information about the truck or driver. In such cases, instead of transmitting all of this data with each message, CDC techniques allows only the changes in the data to be captured, ensuring that the event hub remains up-to-date while minimizing system load \cite{debnath2008sard}. With CDC in place, two streams are consumed: the event stream and the stream of event changes. These streams are joined using a stream-to-static Join, where dynamic event data, such as vehicle positions, is merged with static data, such as driver details. The static data functions as a cache, enabling real-time analysis without requiring redundant information to be transmitted with every event message \cite{van2020evaluation}. Furthermore, to manage the varying data workloads of event data and historical data, streaming and batch processing are often combined, with historical data being stored in a big data platform or an object store, and an event table operating in a pub-sub model, where one subscriber processes the event stream in real-time, and another stores raw data for historical analysis. One example of this is the lambda architecture \cite{lin2017lambda}. Alternatively, one can rely solely on stream processing, e.g. via a Kappa architecture \cite{kreps2014questioning}.

\subsection{AI learning}

As a consequence of the diverse data architectures employed for processing, three distinct AI learning paradigms emerge: (1) offline inference with offline training, (2) online inference with offline training, and (3) online inference with online training.

\textbf{Offline inference / offline training.} The 1\textsuperscript{st} paradigm is typically applied to data at rest, where algorithms process the data multiple times under the assumption that the data samples are drawn from a stationary probability distribution. This assumption allows models to use greedy, hill-climbing search strategies within the model space \cite{kreps2014questioning}. In practice, data scientists often rely on data warehouses for structured data and query data lakes for raw, unstructured data to support offline learning \cite{warren2015big}.

\textbf{Online inference / offline training.} In the 2\textsuperscript{nd} paradigm, model metadata, hyperparameters, vocabulary, and learned weights are periodically analyzed offline and applied using a static inference engine. It is important to note that while data at rest loses value when processed in bulk, data in motion can also lose value, as its patterns are extracted by outdated algorithms with low frequency retraining.

\textbf{Online inference / online training.} The 3\textsuperscript{rd} paradigm addresses this issue by continuously updating the inference engine, eliminating the need for periodic retraining. If such a model could replicate the learning capacity of traditional batch methods, while continuously mining high-volume, open-ended data streams as they arrive \cite{van2020evaluation}, the extracted value from the data would be maximized. However, such guarantees are not feasible, and trade-offs are inevitable. Online training inevitably requires incorporating summaries of prior data, e.g. representative samples or decision models.

\subsection{Network capacity}

In the IoT environment, it is essential to consider not only the selected data processing architectures and learning algorithms but also their deployment locations, and the corresponding network capacity.

\textbf{\textcolor{green}{Edge nodes.}} For instance, remotely located servers connected via the Internet offer significant processing power for managing and analyzing large-scale IoT data. However, these servers can become costly and inefficient when tasked with handling real-time data ingestion at high volumes and strict response time requirements \cite{kreps2014questioning}. The bandwidth required to transmit data to and from the cloud can be substantial, and while predictions can be made online after models are trained offline, real-time processing at scale remains prohibitively expensive, especially in systems with numerous devices and high-speed data streams. Furthermore, high-frequency sampling coupled with bandwidth latency can lead to issues such as data arriving out of order, disrupting the intended sequence of events. To address the challenges of insufficient capacity for transmitting all data to a centralized data center, edge nodes can be deployed closer to the event source. While edge computing is not meant to replace cloud computing, it complements it by enhancing system efficiency, accelerating tasks typically handled by the cloud, and reducing both transmission and computation times.

\subsection{AI resources}

As a result of varying processing locations and hardware choices, the resources available for AI algorithms can differ significantly \cite{schizas2022tinyml}. Machine learning algorithms can generally be categorized into three groups: TinyML algorithms, optimized for battery-operated devices with limited resources, typically around 100KB of RAM and low power consumption; MobileML algorithms, designed for devices with storage capacities up to 8GB, balancing accuracy and efficiency; and CloudML algorithms, which prioritize accuracy over computational constraints, often ignoring hardware limitations \cite{warden2019tinyml}.

\textbf{TinyML.} Table \ref{table:edgedevices} presents four examples of TinyML devices commonly used in IoT applications. As a general rule, devices with clock speeds below 100MHz typically have around 256KB of SRAM, while those with clock speeds above 100MHz tend to have between 256KB and 8MB of SRAM \cite{schizas2022tinyml}. These devices are usually powered by coin or Li-Po batteries, which require power in the milliwatt range \cite{warden2019tinyml}. However, these hardware requirements are constantly evolving as technology advances.

\begin{table}[H]
\centering
\caption{Examples of edge devices for TinyML: a) Himax WE-I Plus HX6537-A, b) Expressif EYE ESP32-D0WD, c) Arduino Nano 33 BLE nRF52840, d) SparkFun Edge ArtemisV1 }
\label{table:edgedevices}
\begin{tabular}{p{0.21cm}p{1cm}p{1.6cm}p{2.71cm}p{0.7cm}}
\toprule
& \textbf{Clock} & \textbf{Flash/RAM} & \textbf{Sensors} & \textbf{Radio} \\
\midrule
a) &  400MHz & 2MB/2MB & Acc, Mic, Cam & None \\ 
b) &  240MHz & 4MB/520kB & Mic, Cam & WiFi \\ 
c) &  64MHz & 1MB/256kB & IMU, Temp, Gest, + & BLE \\ 
d) &  48MHz & 1MB/348kB & Acc, Mic, Cam & BLE \\
\bottomrule
\end{tabular}
\end{table}

\subsection{On-device learning requirements}

Faced with bounded compute and time, an online autonomous system needs to make implicit and explicit tradeoffs, both at a stability-plasticity and a forward-backward transfer level, where the minimum achievable error rate is given by the knowledge provided in the data and hardcoded in the model's inductive biases. Thus, being essential to consider if the assumed constraints are actually useful for the search space at hand. For instance, to design for more restrictive real-life situations, the notion of on-device training is usually associated with overly strong constraints, depicted in Figure \ref{table:batch-stream}, e.g. single pass learning with each instance at a time, and assuming no data storage. While the intention of these requirements may seem logical, defining whether an algorithm should fulfill each one of these constraints implies a trade-off that needs to be analyzed. For example, on-device training may consider incorporating high-frequency mini-batch updating techniques. Additionally, the decision of whether mini-batches are revisited multiple times by the algorithm presents another trade-off. Furthermore, experience replay approaches that keep a small amount of data storage space of previous knowledge, controlling data processing within small batches of instances via curriculum learning, or embedding knowledge on the training algorithm to reduce the plasticity as a bounded data stream nears its end, can all be highly beneficial for on-device training.

\textbf{Natural data flow.} If one must define the constraints of what on-device training is, then the concept of data arrival order and execution model becomes far more critical. On-device ML should operate within a pipelined execution framework allowing data to flow seamlessly through the stages. Regardless of the aforementioned strategies of mini-batching, saving a storage of samples, etc., the primary consideration is that the stages must operate concurrently, with upstream algorithm components actively pushing data downstream as it becomes available. While the aforementioned considerations should matter, this definition brings into focus the core constraint of on-device learning: the inherent lack of control over the arrival of data and corresponding ground truth labels. With data in its natural order, full of temporal, spatial, and causal dependence, and labels that while correct at the time of request may be outdated or even incorrect by the time they become available. It is important, however, not to confuse the issue of processing data in its uncontrolled natural order with the challenge of dealing with non-stationarity. The concept of distribution drift is tied to the expressiveness-convergence trade-off of the model. For instance, while shallow learners may need to fully adapt their representation to incorporate what they interpret as new concepts, deep learners can integrate that same new knowledge with minimal interference. Nonetheless, as deep learners become overwhelmed by data complexity, their convergence rate diminishes, resulting in longer recovery times when adapting to new distributions in an already saturated representation. In contrast, shallow learners may exhibit faster recovery rates, but only if they can effectively represent new concepts \cite{shaker2015recovery}. Ultimately, it is not the data speed that matters, but the interplay between data arrival, training, recovery, and inference. The core challenge lies not in the nature of data itself, but in the computational approach, whether the information is processed immediately as it arrives at each stage, or after being accumulated and processed in earlier stages.

\begin{table}[H]
\centering
\caption{Traditional vs. on-device learning}
\label{table:batch-stream}
\begin{tabular}{ccc}
\toprule
\textbf{Aspect} & \textbf{Traditional} & \textbf{On-device} \\
\midrule
Size & Bounded & Unbounded \\
Processing & Batches & Instance-by-instance \\
Repetition & Multiple & Single pass \\
Data dist. & Stationary & Non-stationary \\
Speed & Slow & Fast \\
Storage & All & None \\
Labels & All & None \\ 
\midrule
Arrival order & Controlled & Natural \\
Execution & Serial, pull data & Pipelined, push data \\
\bottomrule
\end{tabular}
\end{table}

\textbf{Objective evaluation.} Most importantly, to avoid getting caught in the complexities of defining what on-device learning should look like in practice, solutions must be evaluated according to multi-criteria performance metrics tailored to specific edge applications. Intuitively, this can be done by monitoring the learning and retention ability, based on average accuracies and forgetting measures. Moreover, while these output-based metrics are valuable, it is equally important to assess the quality of the representation learned by each algorithm. On-device continual learning is more aptly described as a second-order problem of representation learning, emphasizing how well a model’s internal representations support future learning. Therefore, the focus should not solely be on decision boundaries, but also on how effectively the learned representation facilitates the incorporation of new knowledge. Further details on these evaluation strategies are provided in Section \ref{sec:evaluation}.

\section{Neural networks}
\label{sec:neural}

To address RQ2, this section describes the building blocks needed for deploying NNs at the edge. Firstly, addressing how to embed in NNs the ability to continuously learn, while balancing the stability-plasticity and forward-backward transfer trade-offs. Secondly, covering methods used to squeeze NNs in resource-limited edge devices. Thirdly, adressing how to overcome dependence on external descriptions and heavy supervision in an open world setting. Fourthly, analyzing how well NNs perform on the most dominant IoT data modality, i.e. tabular data, and techniques to improve its performance.

\subsection{Stability-plasticity solutions}

This section covers methods to address catastrophic forgetting, namely controlling how model parameters change between concepts while ensuring independent representations for each concept, capturing common structure within various tasks with an aggregated state abstraction, or decomposing concepts into reusable modules with a compositional representation. Moreover, methods to keep the capacity to learn from new data, as the model accumulates knowledge, are also detailed.

\textbf{Independent representations.} Regarding control of how subsets of model parameters change between concepts, various approaches can be taken. For instance, parameter isolation methods consist in dynamically allocating new neuronal resources when acquiring new knowledge \cite{rusu2016progressive, yoon2018dynamically}. While this strategy directly avoids the problem of catastrophic forgetting, it exacerbates the curse of input dimensionality and storage requirements. As the number of concepts seen grows, there is a larger space of past representations to sift through. Alternatively, one can enforce that different concepts use different components of a single model, typically using attention vectors which protect previous concepts by masking their important parameters and hold-out a subset of its parameters for future learning \cite{serra2018overcoming}. Moreover, some methods use iterative pruning to compact, pick and grow NNs \cite{schwarz2018progress} or initialize context-specific parameters with an auxiliary network \cite{von2019continual}. Intuitively, important parameters should not have their values changed, while non-important parameters are left unconstrained. To create an elastic memory, one can use Fisher Information \cite{kirkpatrick2017overcoming} or KL-Divergence \cite{nguyen2018variational} for weighting in a quadratic parameter difference regularizer of the parameter posterior to control the change of model parameters between two learning tasks. Moreover, multiple disjoint subspaces can be created both at the gradient or feature level. Gradient-based methods impose the parameter update direction of the new concept to be orthogonal to the gradient subspace of the old concepts, ensuring minimal interference. Feature-based methods require that the parameter update direction of the new task is orthogonal to the subspace spanned by the input features of the old concepts \cite{cheung2019superposition, wortsman2020supermasks}.

\textbf{Aggregated state abstraction.} While these approaches mitigate catastrophic forgetting by freezing previous concept representations and minimizing representational overlap, they also minimize the potential for positive forward and backward transfer. Instead, one can tackle forgetting by preserving the input-output relations as the actual object of interest, disregarding how different can be the parameters that produce the same predictive behavior. For instance, reinforcing the importance of old concepts using experience replay, with four key components potentially tuned: (1) rehearsal representation; (2) label strategy; (3) rehearsal policy; and (4) buffer maintenance policy. For the (1) rehearsal representation, one can store raw input data and latent features from hidden layers on a memory buffer \cite{hayes2021remind}, e.g. by replaying compressed representations of activations of the intermediate layers in addition to the input-output pairs \cite{caccia2022new}, or producing these synthetically with a generative model \cite{shin2017continual}. For the (2) label strategy, the memory buffer can hold the true label, which may not always be feasible in real-time applications, or a predicted label by the model. For instance, one can use additional logits distillation, referred to as dark experience replay \cite{buzzega2020dark}. Using these soft targets can be interpreted as functional regularization by generalizing the constraining metric to be a divergence on the output conditional distribution \cite{chaudhry2019tiny}. In contrast to the aforementioned parameter regularization, no assumptions are made on the parametric form of the parameter posterior distribution, which allows models to flexibly adapt their internal representation as needed. For the (3) rehearsal policy, one can identify which stored samples should be selected for rehearsal using a variety of sampling policies, e.g., uniform balanced, min rehearsal, max loss, min margin, min logit-distance, and min confidence \cite{chaudhry2018efficient,prabhu2020gdumb}, or even making virtual updates with the incoming data to find the maximally interfered old samples for a loss-based policy \cite{aljundi2019online}. For the (4) buffer maintenance policy, most techniques use reservoir sampling, where a new sample overwrites a randomly selected sample from the buffer once the buffer is full. However, since the reservoir buffer population approximately follows the data distribution, it can severely deteriorate the performance of underrepresented domains. To alleviate this, one can use a class-based reservoir scheme \cite{de2021continual}, exploit the statistics of the stored samples \cite{prabhu2020gdumb}, or using classification uncertainty and data augmentation \cite{bang2021rainbow}.

\textbf{Compositional representation.} While these methods allow to learn a shared representation, such effort has proven to be equivalent to learning each new concept in a constrained parameter space and NP-hard in general, as the feasible parameter space tends to be narrow and irregular as more concepts are introduced \cite{knoblauch2020optimal}. In fact, these non-modular methods fail to capture the intuition that, in order for knowledge to be maximally reusable, it must capture a self-contained unit that can be composed with similar pieces of knowledge. To alleviate this issue, modular architectures with compositional knowledge allow solving combinatorial previous concepts to learn new ones. There are different types of modularity: data-based, which can be both intrinsic or imposed; concept-based, which relies on sequential and parallel sub-task decomposition; and model-based, which relies on static and conditional composition of sequential or non-sequential modules \cite{sun2020survey}. In practice, modularity is often achieved with dynamic networks, e.g., designed via neural architecture search \cite{liu2018darts}, presenting dynamic depth, width, and routing. For instance, multiple network branches can be parallel built as experts, whose outputs are selectively executed with a gating module and fused with a data-dependent weighting module \cite{shazeer2017outrageously}. However, such solution doesn't fully eliminate weight sharing across modules and instead only gates the sharing. Conversely, routing-level solutions allow for higher specialization, decomposing the pool of sub-concepts in different processing stages or time steps, not only in parallel but also in sequence. Indeed, because of this, modules can be used at multiple locations in the network, and the algorithm can better learn to share parameters dynamically using a composing strategy to choose different parametrized functions depending on the context. Given an input the router makes a routing decision, choosing a function to apply and passing the output back to the router recursively, terminating when a fixed recursion depth is reached \cite{ostapenko2022attention}. Because routing networks jointly train their modules and the router, their major challenge consists in stabilizing the interaction of heterogeneous modules and diverse choices made by the composition strategy training \cite{kirsch2018modular}. In conditional computation, when there is a premature selection of a few modules, which are trained more rapidly, a self-reinforcing effect leads to missing diversity and causes modules collapse \cite{rosenbaum2019routing}. To counter this effect, one can regularize module usage towards a desired form of a sub-optimal minimum \cite{shazeer2017outrageously}. Moreover, this problem extends to the fact that training a routing network is non-stationary from both the perspective of the router, and of the modules, as they mutually depend on each other. Consequently, most methods either fix the composition strategy while learning the modules; or fix the modules while learning the composition strategy. But in its most general form, the compositionality problem should be to jointly learn both \cite{rosenbaum2019routing}.

\textbf{Capacity to learn.} Despite the aforementioned methods tackling the key issue of catastrophic forgetting for on-device training, these didn't consider that, as the model accumulates knowledge, capacity to learn from new data decays. In such situations, one might not want to discard any irrelevant information that is still retained in the model's memory. Naturally, one could simply use larger networks which tend to maintain plasticity longer in face of new concepts \cite{rusu2016progressive,yoon2018dynamically}, however such strategy wasn't designed for a setting of finite memory in edge environments. Instead, plasticity injection lies in a minimalist intervention that increases the network plasticity without changing the number of trainable parameters, by freezing the current model and create a new one which learns the change to the predictions \cite{nikishin2024deep}. Moreover, capacity to learn new data is intrinsically related to maintaining plasticity \cite{lyle2023understanding}. As networks converge to the current concept's local optima, they depart from their initial random distribution, losing the ability to adapt to new learning signals, especially after training for long sequences of concepts \cite{dohare2021continual}. Thus, instead of reactively injecting capacity, one can preventively maintain the plasticity properties of the initial random distribution of network weights via diverse non-saturated units and small weight magnitudes \cite{achille2018critical}. Furthermore, regularization techniques can greatly influence the optimization landscape. For this purpose, instead of regularizing towards zero which is likely to collapse the ranks of the weight matrices, one can regularize toward the initial parameter values \cite{kumar2023maintaining}. Thus, maintaining smaller weight magnitudes, avoiding saturated activations, and preventing weight rank from collapsing.  Alternatively, one can allow the parameters to deviate further from initialization, by instead preserving the curvature properties for desirable plasticity, taking the difference of order statistics using a Wasserstein regularizer \cite{lewandowski2023curvature}. These techniques are similar to the aforementioned constraining of parameters under the umbrella of catastrophic forgetting \cite{kirkpatrick2017overcoming}. However, instead of regularizing towards the parameters of a past concept to preserve old knowledge, these regularize towards initial parameters to remember how to learn. Furthermore, these strategies focus on unit-level interventions, thus potentially missing out on global properties of the loss landscape. To address this, one can regularize a feature subspace to its initial value \cite{lyle2023understanding}. Moreover, one can also shrink all the weights and inject randomness, in order to stop the weight magnitude from continually increasing and reduce the percentage of saturated units \cite{ash2020warm}. However, this approach can be somewhat drastic, causing the model to forget previously learned knowledge. To alleviate this, one can guide this process, by resetting units which have been saturated \cite{sokar2023dormant}, or using more sophisticated approaches. For example, tracking a mean-corrected contribution utility of the outgoing weights and a adaptation utility of incoming weights for each neuron \cite{dohare2021continual}, in order to reset a neuron when the product of these utility measures goes below a certain threshold, while also protecting new units from immediate re-initialization with a maturity threshold. Overall, resetting improves plasticity, but it can slow down convergence on single concepts, while still not resolving the fundamental signal propagation issues of stable representations \cite{lyle2023understanding}.

\subsection{Memory constraints}

Most of these models employ multiple memory systems without considering processing, memory, and time limitations, which severely limits their deployment on resource-limited devices and real-time applications. To circumvent this issue, numerous model compression and acceleration methods have been proposed, both at a data and algorithm level.

\textbf{Algorithm-level.} From the algorithm side, common techniques are: pruning, which lies in removing less significant weights and connections \cite{han2015learning,liu2019rethinking}; quantization, which reduces the precision of the network's weights and activations \cite{courbariaux2015binaryconnect,nagel2019data}; lightweight NN designs, like MobileNets \cite{howard2017mobilenets} and ShuffleNet \cite{zhang2018shufflenet}, whose architectures are specifically crafted to be efficient from the ground up; low-rank factorization which further enhances efficiency by decomposing weight matrices into lower-dimensional forms \cite{jaderberg2014speeding, denton2014exploiting}; and self-distillation in which a NN uses its own predictions to iteratively refine and improve its performance \cite{zhang2021self, zhang2019be}. Regarding pruning, one can employ L0 or L1 regularization to induce sparsity directly \cite{liu2017learning, louizos2017learning}, or identify the connections to prune with a inferred measure, such as the L2 weight magnitude \cite{han2015learning,park2020lookahead}, first-order \cite{molchanov2016pruning} and
second-order \cite{dong2017learning} information, or both \cite{molchanov2019importance}. 
However, most of these methods focus on removing individual weights from the network, resulting in highly irregular sparse models that require specialized hardware and software implementations for efficient execution. To circumvent this issue, one can create faster models by removing entire structures within a network, such as channels, filters, or layers, thus maintaining compatibility with existing hardware accelerators and software libraries \cite{wen2016learning, li2016pruning}. 

\textbf{Data-level.} From the data side, one can try to efficiently condense more information into the memory by sequentially distilling the knowledge from the data, while keeping the model fixed. Thus, these methods optimize the pixel values of distilled images, instead of optimizing the network weights for a particular training objective. For instance, it has been shown that it is possible to compress 60,000 MNIST training images into just 10 synthetic distilled images (one per class) and achieve close to original performance with only a few gradient descent steps, given a fixed network initialization \cite{wang2021dataset}. In fact, these small number of data points, given to the learning algorithm as training data, do not need to come from the correct data distribution in order approximate the model trained on the original data.

\subsection{Supervision assumptions}

Most of these models depend on external descriptions and heavy supervision assumptions, failing when there are unknowns, referred to as the open world setting \cite{salehi2021unified}. For instance, compositional representation methods without access to the information about which part of the input should be used for module selection would perform poorly for samples that are compositionally out of the training distribution. To circumvent this issue, most of these models assume that semantic information is explicitly given in the form of external task descriptors, attribute information, pretrained modules, or hard-coded modular structures \cite{salehi2021unified,lee2022theoretical}. In classification tasks where semantic information is available, it is only necessary to consider known known classes (KKCs) and known unknown classes (KUCs), i.e. positive and negative training samples, respectively. However, in evolving context-dependent scenarios, additional categories must be considered. These include unknown known classes (UKCs), which have limited training samples but are associated with accessible side-information, and unknown unknown classes (UUCs), which lack both training samples and semantic side-information. 

\textbf{UKC.} To tackle a UKC, both generalized few shot learning and zero-shot learning are used. Generalized few-shot learning leverages the relationships between KKCs and a small set of labeled samples resembling the UKC, incorporating shared semantic information across KKCs, UKCs, and the available labeled samples \cite{wang2020generalizing}. In contrast, generalized zero-shot learning relies exclusively on semantic information shared between KKCs and UKCs to recognize novel categories during testing. Here, test samples originate from both KKCs and UKCs, but no labeled instances similar to the UKC in question are provided during training \cite{xian2018zero}.

\textbf{UUC.} However, total autonomy requires models to learn exclusively from the implicit structure inherent to the input data, leveraging internal cues to distinguish between concepts \cite{wang2020generalizing}. Under this paradigm, UUCs can be addressed through open-set recognition and novelty detection methods. Open-set recognition handles UUCs by rejecting them when only KKCs are available, without additional side-information, such as attributes, and when there is a limited number of UUC samples \cite{geng2020recent}. Conversely, it can also recognize UUCs when semantic information shared between KKCs and UUCs is available. Novelty detection methods, on the other hand, address UUCs without relying on any explicit side-information by decomposing the problem into two levels: (1) distinguishing between learned and novel concepts, and (2) incrementally learning and solving each individual concept. For instance, performing cross-concept class discrimination \cite{guo2023dealing} or identifying concepts via predictive and parameter uncertainty, e.g. using conditioned hypernetworks \cite{von2019continual}, or learned binary masks \cite{wortsman2020supermasks}. Moreover, one can explicitly model the cross-concept identification, e.g., using a separate network \cite{abati2020conditional}, a differentiable Bayesian changepoint detection scheme \cite{harrison2019continuous}, or any out-of-distribution detection method \cite{lee2022theoretical}.

\subsection{Tabular data adaptations}

Learning from tabular data forms the backbone of numerous real-world IoT applications, largely due to the wide adoption of relational databases in these domains. NNs have enabled remarkable advancements in handling unstructured data, such as text and images. However, when applied to tabular data, NNs often underperform DT ensembles \cite{borisov2022deep}.

\textbf{Tabular challenges.} As deep learning architectures have been crafted to create inductive biases for homogeneous data, matching invariances and spatial dependencies of the data, they struggle to learn irregular patterns of the target function, with their rotation invariance hurting performance \cite{grinsztajn2022tree}. While in text and vision, data is intrinsically tied to the position of the word/token or pixel, the order-invariance of tabular data hinders position-based methodologies, e.g. convolutions, less applicable for tabular data modeling. This lack of spatial or temporal structure that could inform a good prior of the data makes it challenging for the model to understand the inherent relationships between features. Moreover, this is further aggravated when handling heterogeneous and uninformative features, skewed heavy-tailed feature distributions, and extreme values \cite{grinsztajn2022tree}. NNs trained on tabular data tend to develop overly simple decision boundaries, biased to overly smooth solutions and low-frequency functions. In contrast to extensive preprocessing required for NNs, DTs handle these issues effectively by approximating missing values, using appropriate thresholds and managing variable ranges internally during the splitting process \cite{shwartz2022tabular}. To circumvent these issues, various techniques have been proposed to make deep learning work on tabular data, namely through data transformations, regularization models, differential trees and transformer-based models \cite{borisov2022deep}. 

\textbf{Data transformations.} Firstly, one can convert the tabular input into a homogeneous data format, e.g. by transforming tabular data into images for direct application in a convolutional NN \cite{tan2020converting}. Moreover, for numerical features, embeddings have shown to provide powerful initial representations, e.g. replacing the original values with bins indices constructed with quantiles of the individual feature distributions or the target values, using Johnson distribution or piecewise linear transformations \cite{gorishniy2022embeddings}. Alternatively, one can use more complicated automatic techniques that are part of the model architecture, e.g. using a contextual embedding with self-supervised learning \cite{yoon2020vime}.

\textbf{Regularization models.} Alternatively, one can model the varying importance of tabular features and the multiplicative interactions between them. For instance, by using gradient attributions to encourage neuron specialization and orthogonalization \cite{jeffares2023tangos}, a counterfactual loss \cite{katzir2021net}, an asymmetric encoder-decoder self-supervised framework \cite{wu2024switchtab}, constructing and projecting into a prototype-based space \cite{ye2024ptarl}, or stochastic gates \cite{yamada2020feature}.

\textbf{Differential trees.} To benefit from the inductive biases of tree-based methods, these can be extended with a smoothed decision function. For instance, by combining gradient-based optimization with hierarchical representation learning \cite{popov2019neural}, using a gradient boosting framework with shallow NNs as weak learners \cite{badirli2020gradient}, using tree-inspired sequential attention for feature selection \cite{tabnet2019arik} or distilling a NN into a soft DT \cite{frosst2017distilling}.

\textbf{Transformers.} Alternatively, one can use token-based mechanisms for feature selection and reasoning in heterogeneous tabular data \cite{shwartz2022tabular}. For instance, combining column descriptions and cells as input for feature encoding \cite{wang2022transtab}, row and column attention mechanisms to capture the inter-sample interactions with self-supervised pretraining to deal with label scarcity \cite{somepalli2021saint}, additive attention to consider the interaction between each token and the global representation, achieving a linear complexity \cite{wu2021fastformer}, transforming categorical features into contextual embeddings \cite{huang2020tabtransformer}, or sequential hierarchical subnetworks prioritizing the most significant features \cite{tabnet2019arik}. Furthermore, some of the aforementioned tabular transformers \cite{wu2021fastformer, somepalli2021saint}, when pre-trained with a diverse range of tabular datasets, have been shown to exhibit better performance via finetuning than training from scratch \cite{zhu2023xtab}. Following these findings, various large tabular models pretrained on synthetic tabular data have been proposed for classification, which yield great promise for application on edge applications \cite{hollmann2025accurate, bonet2024hyperfast}. By using on-the-fly sketches to summarize unbounded streaming data, one can feed this information into the pre-trained foundational model for efficient processing \cite{lourenco2025incontext}.

\section{Decision trees}
\label{sec:trees}

To address RQ3, this section describes the adaptations needed to make DTs viable for on-device training at the edge. Firstly, addressing how to dynamically adjust the tree structure to new data, via approximation-based splitting. Secondly, describing strategies that allow to control the tree growth efficiency in memory-scarce edge environments. Thirdly, dynamic combinations thereof in ensembles for better performance, including meta-learning and model-based clustering techniques to model a concept history. Fourthly, how to overcome supervision assumptions. In reviewing these methods, it is important to note that while NNs can equally evolve via both dynamic architectures, i.e. adding new parameters, and fixed architectures, i.e. activating the relevant part of the network without changing the architecture, the algorithm properties of DTs put more emphasis on adaptation via dynamic architectures, i.e. structural expansion and pruning.

\subsection{Incremental learning}

To incrementally construct a DT at edge devices, using contained memory and time per sample, the fundamental algorithm design component is approximation-based splitting \cite{domingos2001catching}. As new instances arrive, they traverse the tree from the root to a leaf node, updating statistics at each node that guarantee for split decisions almost identical to those that conventional batch learning methods would make, despite lookahead and stability issues. 

\textbf{Split decision.} Based on these sufficient statistics, these DTs continuously update the heuristic values for each attribute, such as the traditional information gain (IG) and Gini index (GI) \cite{domingos2000mining}, fuzzy entropy \cite{ducange2021fuzzy} or a fair-enhancing gain \cite{zhang2020feat}. With this, the decision to split is built upon the concept of a statistical Bound which quantifies the confidence interval for the heuristic function, based on a minimum amount of data needed to make a statistically reliable decision when the data distribution is unknown. For instance, comparing the Hoeffding Bound (HB) to the difference of heuristic obtained between the best attribute split and second-best attribute \cite{domingos2000mining}. Once the difference in evaluation surpasses the bound, the leaf node is split into child nodes. However, this HB can be too restrictive and its expected guarantees not given. Firstly, it only applies to numerical averages, however both IG and GI cannot be expressed as arithmetic average. Secondly, it demands that these objects are independent variables, with the observations used for the computation of the split criterion being statistically independent. In most cases, the assumption of independence is not valid, primarily due to the overlapping nature of sliding windows and the inherent sequential characteristics of data streams. Alternatively, one can directly use the classification error as a splitting criterion, rather than a concave approximation of it, like the entropy or the Gini index \cite{rutkowski2014new}, decompose the entropy gain calculation into three components and applied the HB to each one of them \cite{duda2014novel}, or use the McDiarmid's bound, which provides tighter guarantees in situations where the impact of changing one variable is limited by a known constant, being tailored for functions of independent random variables with bounded differences \cite{rutkowski2012decision, de2017confidence}. As for the leaf classification strategy, despite most algorithms using majority voting (MV), VFDTc has shown that one can apply the naturally incremental naive Bayes (NB) classification in tree leaves for better exploitation of the available information \cite{gama2003accurate}. While VFDT only uses a crude approximation about class distributions, NB takes into account not only the prior distribution of the classes, but also the conditional probabilities of the attribute-values given the class. Alternatively, one can also dynamically switch between MV and NB, depending on the amount of data available.

\textbf{Split point.} Another challenge is finding the split value. For categorical attributes, memory requirements are relatively low because the algorithm only needs to store the frequency counts of each unique value \cite{domingos2000mining}. For continuous attributes, however, it is impossible to evaluate all possible split points to determine the optimal choice, and on-device training prohibits knowing the range of values a priori, thus inhibiting proper discretization \cite{gama2006discretization}. To address this issue, the original VFDT used a binning approach where for every new unique value observed a new bin is inserted into the range. A single set of bins is shared by all of the classes. Once a predefined limit of 1000 bins is reached, the bins remain stationary, but their counts are incremented. Thus, not only being influenced by just the first 1000 unique values, but also requiring a lot of processing time and memory usage \cite{domingos2000mining}. VDFTc mantains a binary search tree at each leaf node, dynamically inserting new values, and tracking class counts on either side of each cut point \cite{gama2003accurate}. Alternatively, the distribution of values can be approximated by a single Gaussian \cite{gama2004forest}, with the range between the minimum and maximum observed values being divided equally into \(n\) parts. For each of these split possibilities, the distribution of classes to the left and right is estimated based on the area under the normal distributions \cite{holmes2005tie}. Thus, requiring the update of a few counts per class. Moreover, it has been shown analytically that by using Gaussian distribution \cite{holmes2005tie} as sufficient statistics and misclassification error \cite{rutkowski2014new} as impurity measure, the best split value is where two Gaussian distributions intersect each other \cite{mirkhan2021analytical}.

\subsection{Dynamic pruning}

The aforementioned choices allow to produce increasingly larger trees that can divide the problem space into finer regions, better approximating the target concept. However, such tree growth still remains highly uncontrollable, which can lead to two main issues: excessive memory consumption due to multiple redundant splits on all features, and a loss of plasticity due to the descendant nodes getting subsequently fixed to the space covered by their parent node. To address this, various pre-pruning and post-pruning techniques have been proposed. Pre-pruning strategies include splitting rules enhancements, adaptive tie breaking, adaptive grace periods, activity-based node decisions, and removing poor attributes. Post-pruning includes various methods that estimate in each decision node whether the corresponding rooted subtree is consistent with the current data, then performing some action to reconstruct the subtree if an inconsistency is estimated.

\textbf{Split rules.} Traditional incremental DTs primarily evaluate the top two attributes for splitting. However, additional splitting criteria can be introduced to enhance the quality of splits. For example, constraints can be implemented to compare current metrics with both historical and cross-leaf data, while a skipping mechanism can be used to bypass strict splitting conditions when significant data changes are detected \cite{da2018strict}. Another approach involves monitoring the fluctuation of the HB by tracking its mean, minimum, and maximum values, along with accuracy metrics, to serve as pre-pruning conditions for splits \cite{yang2011optimized}. Additionally, a constraint can be incorporated to assess whether the top-ranked splitting feature at a leaf node provides substantial information gain compared to previous splits along the same branch \cite{barddal2020regularized}.

\textbf{Tie breaking.} When two attributes have statistically similar IG, tree expansion can stall, even if either attribute would be a viable splitting choice. To avoid this, a split is triggered when the bound falls below a predefined threshold, ensuring a minimum growth rate. However, this fixed threshold can lead to premature splits, not due to a meaningful tie, but because of a lack of strong candidates. To address this limitation, an adaptive approach can be used, where a threshold is dynamically adjusted based on the mean of the HB, leveraging its proportionality to the number of input stream samples \cite{yang2011moderated}. Another alternative is to determine tie-breaking conditions by comparing the difference between the top two candidates with the difference between the second-best and worst candidates, providing a more context-aware decision-making process \cite{holmes2005tie}.

\textbf{Split attempts.} To prevent premature splits due to insufficient sample sizes, which can compromise the reliability of the HB, decision trees impose a grace period, requiring a minimum number of observed instances before a split is considered. However, a fixed grace period that does not adapt to the data being processed can lead to inefficient computations, either by triggering unnecessary split attempts or by introducing delays in tree growth. Instead, a more effective approach is to adjust the waiting period dynamically based on observed tie-breaking patterns \cite{holmes2005tie}. For instance, being adjusted dynamically after an unsuccessful split attempt, ensuring that the next iteration leads to a valid split. If the top-ranked attributes have somewhat distinct IG values, but not enough to justify a split, the algorithm can increase the grace period so that more samples are gathered until the HB falls below the IG difference. Conversely, if the best attributes are highly similar in IG but do not surpass the tie-breaking threshold, additional instances are required to reduce the HB below the threshold, with the grace period adjusted accordingly \cite{garcia2018hoeffding}. Alternatively, one can analyze class distributions from past split attempts to estimate the number of additional instances required for the HB condition to be satisfied \cite{losing2018enhancing}. To bypass this split attempt problem, a further optimization involves reducing the computational cost of split evaluations by replacing periodic heuristic recalculations with an incrementally updated heuristic measure, which can be maintained in constant time \cite{sun2021speeding}.

\textbf{Leaf activity.} Another pre-pruning strategy focuses on regulating split decisions based on leaf activity. The likelihood of selecting a suboptimal split depends not only on the number of examples within a leaf but also on factors such as feature space dimensionality, leaf depth, and the total number of instances processed by the algorithm \cite{de2017confidence}. One way to address this is by comparing the number of samples that reach existing leaves to those that do not \cite{yang2011optimized}. If most samples continue to fall into existing leaves, tree updates can be delayed to prevent unnecessary structural modifications. Alternatively, leaves with low utility can be deactivated based on their probability of receiving new examples and their observed error rate \cite{domingos2000mining}. When a better split is identified, the previous split’s leaves are deactivated, and their statistics are stored. If, during a later evaluation, a deactivated split is again determined to be optimal, its saved statistics can be restored, avoiding the need to restart the process. A more adaptive approach normalizes leaf activity by measuring how many instances a leaf has processed relative to the average across all leaves since its creation \cite{garcia2022green}. Depending on this value, different strategies can be applied: inactive nodes can be halted to conserve resources, while high-activity nodes can trigger more aggressive expansion techniques, such as skipping constraints in the splitting process \cite{lourenco2025dfdt} or switching to a more flexible splitting rule \cite{garcia2022green}. These adaptive mechanisms ensure that computational resources are focused on expanding the most relevant nodes while preventing unnecessary splits in less significant areas of the tree.

\textbf{Attribute removal.} Finally, one can selectively discard leaf statistics for attributes that are likely to not cause a split. For instance, by evaluating after an unsuccessful split attempt, if the attribute's IG is less than the gain of the current best attribute by more than the HB \cite{domingos2000mining}. To further reduce computational cost, one can introduce a dynamic candidate attribute set which keeps track of the top K attributes that are most likely to split. Only the heuristic measures of attributes in the set are updated continuously \cite{sun2021speeding}. Also, one can complement both strategies, eliminating poor attributes that increase computational complexity, while grouping candidate good attributes to be used in the next iteration, instead of using all of the attributes \cite{lefa2022enhancement}. Another perspective on this issue is to prevent from using features that have not been used thus far, by penalizing those that have similar gains to other features in previous splits \cite{barddal2020regularized}. Naturally, while the aforementioned pre-pruning methods allows to contain the model size, they are still unbounded. To address this, any of the aforementioned strategies can be guided by memory limits \cite{domingos2000mining}. To avoid the costly check of the current memory, one can perform the precise calculation periodically, while estimating the size approximately whenever new nodes are introduced to the tree.

\textbf{Split re-evaluation.} Regarding post-pruning, methods can be characterized by two key steps: estimating in each decision node whether or not the subtree rooted at this node is consistent with the current data, and performing some action to reconstruct that subtree if inconsistent. To estimate consistency, one can re-evaluate every split in the tree to ensure that the current split remains the optimal one, e.g. checking if the current split outperforms the best alternative \cite{hulten2001mining} or the null attribute \cite{ manapragada2018extremely}. Alternatively, one can directly monitor the error rate in nodes \cite{ikonomovska2011learning}. For instance, computing the error top-down with the prediction from the current split node \cite{gama2004forest}. Or, instead, computing it bottom-up with the prediction from leaf nodes \cite{bifet2009adaptive,losing2018enhancing}. For this purpose, most approaches monitor streaming bits of misclassification \cite{gama2005learning,bifet2009adaptive}. However, one can also monitor performance as a stream of real values, by mantaining a sliding window with relevant instances at each leaf node \cite{nunez2007learning}. Both selectively forgetting outdated instances when leaves are in an improvement state, and aggressively shrinking window size when a persistent degradation of the performance measure is estimated \cite{nunez2007learning}.

\textbf{Tree reconstruction.} To reconstruct the tree, one can simply require another attribute to become the top split attribute \cite{manapragada2018extremely}. If an improved split is detected, a new leaf node replaces the entire subtree rooted at the faulty node. However, while this split revision procedure can significantly reduce variance by simplifying the model, it may introduce considerable bias if important information is lost when the subtree is pruned. To compensate this, one can make new branches more readily, when it has evidence that a specific branch is better than none, i.e. a split with greater-than-zero IG with high probability, rather than better than any potential alternative \cite{manapragada2018extremely}. In other words, instead of finding a split attribute that will never have to be replaced, the model aims at finding the current best available split attribute at a node until a better one is found. Alternatively, one can overcome this by instead restructuring the otherwise pruned subtree, leveraging on the fact that the order in which an instance traverses the nodes of a DT from root to leaf does not affect prediction 
\cite{heyden2024leveraging}. Also, one can hold a short-term memory to train an alternative subtree that could quickly replace the corresponding original one \cite{gama2005learning, nunez2007learning}. However, this strategy is only effective when the underlying change is abrupt, as the old classifier is immediately discarded. In real situations, changes are often gradual and consecutive concepts are similar. Therefore, an old subtree trained with many examples can be more accurate than a new subtree trained with a few, even when the performance of the old subtree has dropped significantly. To be more conservative, it is necessary to determine whether the alternative subtree is more coherent with respect to the current concept, and subsequently decide when to eliminate the alternative subtree that is no longer effective. For instance, building an alternative subtree as soon as an inconsistent node is detected, which only eventually replaces the original subtree, after a user-defined number of examples, when its accuracy is superior to the original subtree \cite{hulten2001mining}. This alternate tree reduces bias while keeping the model relevant, ensuring that the new subtree adapts to the new concept. Moreover, one can perform this adaptation recursively in each alternative subtree affected \cite{losing2018enhancing}, and even create new alternative subtrees in a split node regardless of a previous alternative subtree still being induced by a previous drift signal \cite{bifet2009adaptive}. To leverage on the existence of these various alternatives, one can combine their predictions with a weighted vote approach inversely proportional to the error rates calculated by the change detectors \cite{losing2018enhancing}.

\textbf{Change detection.} These tree reconstruction strategies impose two key things to monitor. Firstly, the prediction errors between subtrees, e.g. using fading factors with the Q statistic \cite{gama2013evaluating} in a user predetermined evaluation interval \cite{ikonomovska2011learning}, or with a predictive sequential approach \cite{losing2018enhancing}. Secondly, the error of the trees susceptible to be pruned, e.g. using the Page–Hinckley \cite{mouss2004test} which is designed to detect abrupt changes in the average of a Gaussian signal \cite{ikonomovska2011learning}, the adwin change detector \cite{bifet2009adaptive} which provides mathematical guarantees for the false positive and false negative, or the hddmA test \cite{frias2014online}, which not only provides the guarantees, but also has constant time and space complexity \cite{losing2018enhancing}.

\subsection{Concept history}

While single DTs offer simplicity, when memory allows, ensemble strategies can be another effective approach. Upon a concept drift, the diversity among trees can be exploited as some local minima are less affected than others. Affected components can be dynamically deleted, added, and combined with a voting scheme, weighting different weak learners based on their past performance. For instance, incorporating incremental versions of random forests \cite{gomes2017adaptive}, bagging \cite{oza2001experimental} and boosting \cite{gunasekara2024gradient}.

\textbf{Pool.} However, such ensembles constantly train all active base classifiers, progressively discarding some of them when a drift is detected. Base learners evolve and the previously learned concepts are forgotten before they reoccur. Consequently, these models might need to be trained from scratch, which results in a waste of computational resources, longer training times, and significant prediction errors while models are not up-to-date with the latest state of the data stream. To circumvent this issue, meta-learning \cite{anderson2016recurring} and model-based clustering \cite{xu2016mining} allow to more explicitly retain a pool of previous active and inactive concepts. This feature enables storing cold copies of previously learned base classifiers to be reused if they become relevant again in the future.

\textbf{Recognition.} Consequently, maintaining a concept history creates the need of identifying what concept is present at each time. For this purpose, there are 2 major categories of techniques acting as a trigger: conceptual equivalence and conceptual similarity. The former assumes that two classifiers describe the same concept when predicting similarly during a time window \cite{anderson2016recurring}, while being enriched with automatic ensemble size calculations \cite{pietruczuk2017automatic}, and time dependencies modeled according to the concept drift properties \cite{almeida2018adapting}. The latter recognizes similar concepts based on cohesion and separability in the input space, e.g. euclidean distances between clusters representing different concept clusters \cite{xu2016mining}. In these, micro-clusters or latent features are used to make a synopsis of the incoming instances and reduce the computational cost of finding similarities among conceptual vectors \cite{katakis2010tracking}. Then, methods differ in how they manage this concept history.

\subsection{Lack of supervision}

Across these methods, adaptation is mostly achieved via monitoring the standardized absolute error of model predictions, leveraging on the fact that changes in the decision boundary manifest in the error functions of learning algorithms as changing optima \cite{mouss2004test}.

\textbf{Unsupervised.} To overcome the reliance on supervision, novelty detection methods use a concept history framework that learns the characteristics of normal patterns, enabling them to proactively address novel patterns in an unsupervised, dynamic environment where concepts may evolve either gradually or abruptly \cite{faria2016minas}. These methods employ clustering-based algorithms to distinguish sparse, noisy examples from distinct, cohesive concept types, by forming closed boundaries around normal patterns. The underlying assumption is that samples close in the input or latent space originate from the same concept. Evolving groupings are identified when sufficient cohesion is present within clusters, with separability allowing for differentiation between new groupings and established concepts or classes \cite{faria2016minas,spinosa2009olindda}.

\textbf{Semi-supervised.} However, while these learning algorithms have the advantage of requiring only unlabeled data, it is important to maintain the ability to leverage any available labels to better define decision boundaries. For instance, it has been shown that semi-supervised anomaly detection methods can outperform unsupervised methods even with only 1\% of anomalous instances \cite{han2022adbench}. In real-world applications, it is easy to imagine a scenario where at least a few labels are available in deployment. Moreover, at the beginning of the learning process, models usually have access to a lot of labeled data. In these situations, one can couple the state-of-the-art performance of supervised learning algorithms \cite{gomes2017adaptive, gunasekara2024gradient} with unsupervised mechanisms used to update the model during execution. For instance, using an ensemble of classifiers to maintain a knowledge base of previous concepts, where instances falling inside the decision boundary of the ensemble are classified by taking the majority of the votes in the ensemble, while the others are queued in a buffer, clustered using k-means and periodically tested for cohesion and separability \cite{masud2011classification}. Alternatively, using active learning mechanisms, e.g. based on instances being an outlier or the proportion of majority votes \cite{masud2013classification}, while triggering label requests based on drifting classifier's confidence scores \cite{haque2016sand}.

\textbf{Change detection.} Indeed, instead of using solely loss-based drift detectors to update state-of-the-art supervised learning algorithms \cite{gomes2017adaptive,gunasekara2024gradient}, one can extend these to deal with scarcely labeled streams. For instance, making the classifier more robust to extreme verification latency via a double-stage detector, which applies CUSUM for change detection and, then, silhouette detection for cluster validation of new classes \cite{costa2019novelty}. Another option is to apply statistical tests on distributional changes in the input space, for instance, by using specialized committees of incremental Kolmogorov-Smirnov tests \cite{dos2016fast}, or a discriminative classifier \cite{gozuaccik2021concept}. However, as these data-based drift detectors totally exclude the classifier from the detection process, this results in increased sensitivity to change and a large number of generated false alarms. In the extreme, if the underlying marginal data distribution $P(X)$ over the input does not contain information about changes in $P(Y\mid X)$, then it is impossible to exploit unlabeled data to improve a supervised learner or detect a change \cite{vzliobaite2010change}. To circumvent these issues, rather than solely monitoring drifts through statistical tests on the data distribution $P(X)$, emerging drift detection methods focus on monitoring proxies of the conditional classifier probability $P(Y \mid X)$ in an unsupervised way. Examples include comparing the distribution of classifier output confidences \cite{lindstrom2013drift}, or tracking the proportion of samples within the margin of a linear support vector machine \cite{sethi2016md3}.

\section{Evaluation}
\label{sec:evaluation}

To address RQ4, one must firstly assess the suitability of the on-device training strategy according to the available relevant data, and hardware investments, relying on the notions discussed in Section \ref{sec:edge} and other criteria, such as privacy and security needs. For instance, distinguishing between fast streams, e.g. in industrial machinery, which require immediate processing or even discarding instances to avoid potential data overload, and slow streams, e.g. in medical applications, which allow for more thorough data processing, and temporarily storing data for later use during periods of lower activity. 

\begin{figure}[h]
    \centering
    \includegraphics[width=\linewidth]{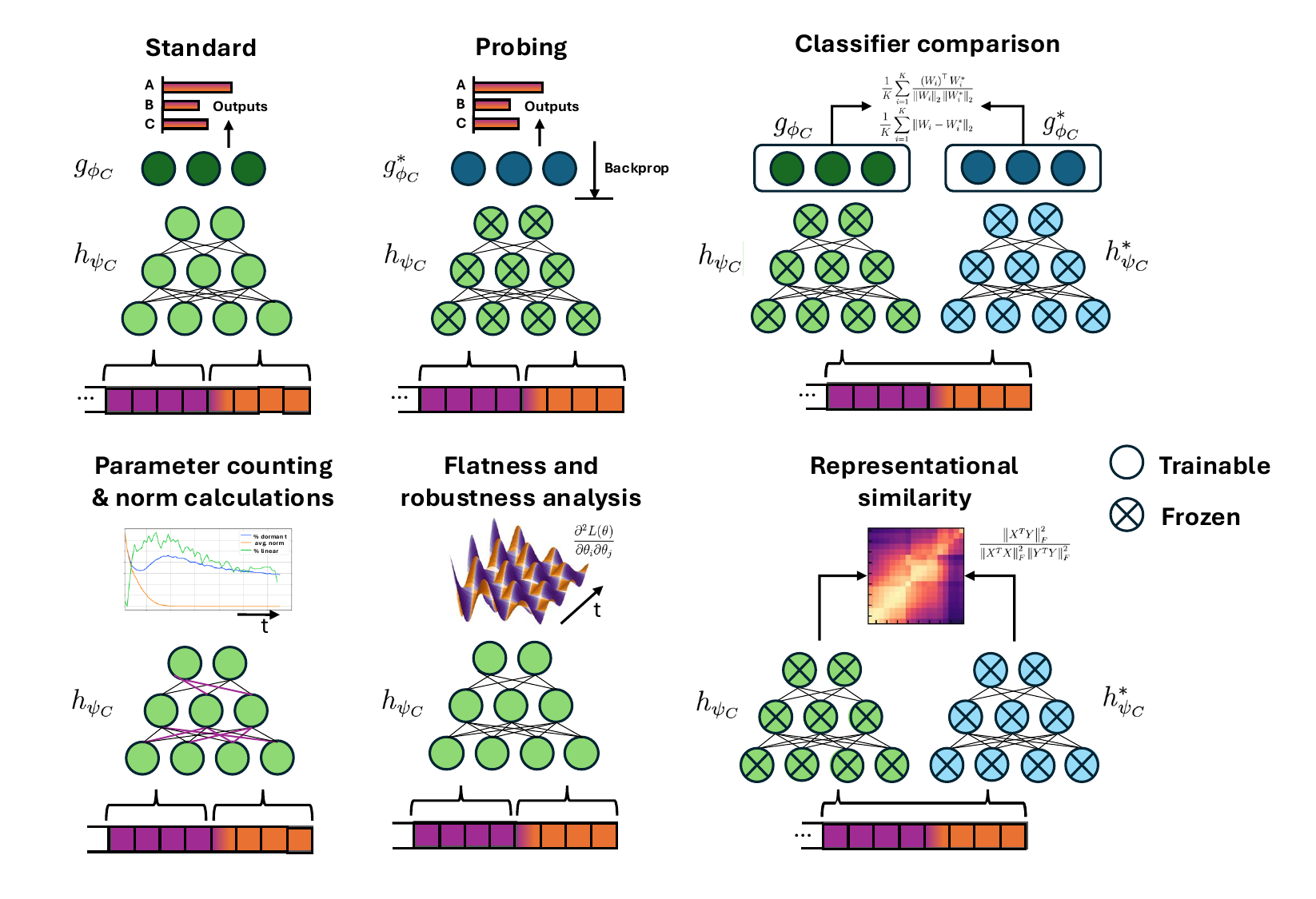}
    \caption{Methods of evaluation: standard, parameter counting or norm calculations, flatness and robustness analysis, probing, representational similarity, and classifier comparison}
    \label{fig:evaluationmethods}
\end{figure}

\textbf{Representation vs data.} After properly formulating the problem, one can focus on accurate multi-objective evaluation to study how different algorithms perform before deployment, both at a representational- and data-level \cite{cha2022towards}. These metrics not only assist in the development of algorithms at a more fundamental level, but also in the creation of better benchmarks, and the a priori estimation of the cost and limits of running learning algorithms in specific edge applications. From the representational-side, one can explore the value of hidden layers, e.g. by retraining the network while freezing layers, resetting blocks of layers to initial values, and re-randomizing parameters \cite{zhang2023layers}. From the data-side, one can interpolate similarity in the data with linear combinations of pairs of instances from different concepts to probe continual learning performance effects \cite{ramasesh2021anatomy}, or study how hardness, complexity, heterogeneity and dissimilarity of the concept sequence can influence learning over time \cite{kemker2018measuring}. In this section, we bring attention to six different groups of techniques, illustrated in Figure \ref{fig:evaluationmethods}: standard evaluation on unseen drifting data, parameter counting or norm calculations, flatness and robustness analysis, probing, representational similarity, and classifier comparison.

\textbf{Standard evaluation.} Common metrics are the accuracy, learning, and forgetting measures  \cite{chaudhry2018efficient} using a per-iteration scheme with a window of size $w$. For instance, defining the windowed plasticity as the maximal accuracy increase in a iterations window, and the stability-based windowed forgetting as the maximal accuracy decrease. By varying this window length, one can account for forgetting and remembering capabilities at different time scales. Nevertheless, despite useful, these metrics don’t provide a view on the maximum achievable performance. For this purpose, one can define the performance in comparison to baseline classifiers. One approach is to consider a model trained on all concepts jointly rather than sequentially \cite{cha2022towards}. However, this doesn’t isolate forgetting effects, since concurrent training can improve the representation for each concept itself by contrasting a single class with a wider variety of other classes. Instead, one can use an ensemble strategy where training remains concept-per-concept and a model copy is stored after every concept \cite{hess2023knowledge}. This allows to reliably observe forgetting effects since the concept’s original representation remains intact and they can all be simultaneously recovered by a concatenated head. Naturally, these upper bound baselines either require the whole data upfront or linearly more compute with every concept. For drift-related performance metrics \cite{gama2013evaluating}, one can also consider: the mean time between false alarms (MTFA); the false alarm rate defined as $1/\text{MTFA}$; the missed detection rate (MDR), the mean time to detection (MTD), and the average run length ($\text{ARL}_\theta$) which generalizes over MTFA and MTD to quantify how long to wait before a change detection of size $\theta$ in the performance. For this purpose the following convention is used: before true change happens, all the alarms are considered as false alarms; after a true change occurs, the first detection that is flagged is considered as the true alarm; after that and before a new true change occurs, the consequent detections are considered as false alarms; if no detection is flagged between two true changes, then it is considered a missed detection.

\textbf{Parameter count and norm.} Regarding parameter counting, the simplest explanation for loss of plasticity is that the gradient norm goes to zero, eliminating all the network’s expressivity. In practice, one can monitor the number of dormant units, i.e. neurons with ReLU’s output being zero or sigmoid’s and tanh’s outputs being too close to either of the extreme values of the activation function \cite{sokar2023dormant}. Thus, not propagating gradients back to its incoming weights and severely slowing down learning. As for norm calculations, a steady growth of the network’s average weight magnitude has been associated with plasticity loss, suggesting a slow divergence in the training dynamics, exploding gradients and numerical overflow errors \cite{mirzadeh2020understanding}. Consequently, this can have a saturating effect on model components, e.g. softmax attention heads \cite{liu2023small} and normalization layers \cite{merrill2023effects}. Thus, resulting, for a fixed update step size, in smaller output changes than those obtained at initialization. Moreover, a more subtle issue can also arise when the norms grow at different rates across layers, leading to learning instabilities, due to updates to each layer having differential effects on the network output.

\textbf{Flatness and robustness.} While simple to compute, parameter counting and norm calculations can be too coarse in studying both sensitivity and generalization error of changing parameters \cite{lyle2024disentangling}. Alternatively, flatness and robustness analysis of trained models allow to better consider the respective network architectures, helping isolate parameter changes in the objective, gradient and Hessian from changing data \cite{zhang2023layers}. A steady decrease in the curvature of the loss function has been linked to plasticity loss. To define a measure of curvature, there are different available functions of the Hessian matrix of the objective function \cite{mirzadeh2020understanding}. One common measure is the sharpness of local minima, given by the maximum eigenvalue of the Hessian which describes the steepest curvature \cite{keskar2017large}. However, this sharpness measure is coarse-grained, only giving the magnitude of the vector of maximal curvature, and failing to characterize other directions. Another measure is the drop of effective rank of the Hessian matrix, which counts the effective number of directions in the parameter space that meaningfully influence the curvature of the loss function \cite{mirzadeh2023directions}. Similar to the rank of a matrix representing the number of linearly independent dimensions, the effective rank takes into consideration how each dimension influences the transformation induced by a matrix \cite{roy2007effective}, where a low value implies that the information in most of the dimensions is close to being redundant. Over time, the learning algorithm favors low-rank solutions for the current concept \cite{razin2020implicit}, which then serves as the initialization for the next concept, with the effective rank of the representation layer incrementally decreasing. Consequently, this limits the number of solutions that the NN is able to explore at the beginning of a new concept \cite{mirzadeh2023directions}. Furthermore, one can reconcile the geometry of the loss function based on the interpretation of the diagonal entries of the Fisher information matrix of the Hessian rank as an approximation of high local curvature \cite{achille2018critical}.

\textbf{Probing, similarity and classifier comparison.} While these flatness and robustness analyses are useful, it is extremely hard to reliably characterize the optimization landscape, since it can also change as the underlying data distribution changes, even though parameters maintain the same. To alleviate this issue, probing, e.g. freezing the encoder representation and retraining the final linear layer using linear classifiers or k-NN classification, allows for better control \cite{cha2022towards}. Moreover, one can use representational similarity algorithms to directly compare changes in representations during a concept drift. In this case, the major challenge is overcoming the natural nonalignment between hidden neurons across different NNs \cite{raghu2017svcca}. For this purpose, if assuming invariance to orthogonal transformation and isotropic scaling, one can use centered kernel alignment \cite{kornblith2019similarity}. Regarding the classifier comparison, one can naively perform linear classification trained on the activations of two classifiers both before and after forgetting to measure the performance that can be recovered \cite{ramasesh2020anatomy}. Alternatively, one can compare how two classifiers partition the feature space using both high cosine similarity and low \( L_2 \) distance as indicators of similarity \cite{cha2022towards}.

\textbf{Resource efficiency.} Beyond latent representations and predictive performance, it is important to measure the computational resource trade-offs, e.g. using the NetScore metric \cite{wong2019netscore} or RAM-Hours \cite{gama2013evaluating}, the time for both inference and training, and even the energy consumption rate of the algorithm, e.g. accounting for the static leakage power consumed when there is no circuit activity, and the dynamic power dissipated by the circuit, from charging and discharging the capacitor \cite{garcia2018hoeffding}. If one estimates the total execution time as the clock cycle time of the processor multiplied by the clock cycles per instruction and the number of instructions, the resulting energy consumed becomes a function of the energy of each instruction, which can be estimated by analyzing the memory accesses delays, computation requirements (integer and floating point operations), cache accesses, and cache misses \cite{garcia2018hoeffding}.

\section{Conclusion}
\label{sec:conclusion}

In this survey, we explored state-of-the-art methods for on-device training, particularly in the context of classification tasks using neural networks (NNs) and decision trees (DTs). \textbf{RQ1} highlighted that the key constraints for on-device training stem from the data architecture (batch vs. stream) and network capacity (cloud vs. edge). These factors influence algorithm design, particularly in terms of data arrival order and execution models. We also identified a clear need for performance evaluation based on multi-criteria metrics, tailored to specific edge applications.
\textbf{RQ2} examined the building blocks necessary for deploying NNs on edge devices, addressing not only techniques for compressing NNs to fit resource-limited devices, but also challenges such as catastrophic forgetting and data inefficiency. Additionally, we discussed the difficulty NNs face in capturing irregular patterns found in IoT tabular data and operating in open-world settings with minimal supervision.
\textbf{RQ3} focused on the adaptation of decision trees for on-device training, emphasizing memory-efficient methods, overcoming reliance on heavy supervision, and addressing the issue of fixed sub-spaces as descendant nodes evolve. While NNs can adapt through both dynamic (parameter addition) and fixed (relevant parts activated) architectures, DTs rely more heavily on dynamic adaptations such as structural expansion, pruning, and meta-learning to model the concept history.
\textbf{RQ4} emphasized that evaluating on-device training strategies requires more than just output-based metrics. A comprehensive evaluation must also account for the model’s internal representation quality, assessing how well it supports future learning. Multi-objective metrics, such as probing, robustness analysis, and representational similarity, are crucial for understanding how different algorithms adapt to the varied demands of edge scenarios, considering both hardware limitations and data characteristics.

On-device edge training remains a complex challenge. While NNs show great potential, they require significant optimization to address issues such as resource constraints, catastrophic forgetting, and handling tabular data. In contrast, DTs are more efficient and better suited for resource-limited environments, but their limited expressiveness restricts their ability to capture complex patterns. This creates a trade-off between expressiveness and convergence, where DTs must fully adapt to new concepts, while NNs can integrate new knowledge with less interference. However, as data complexity increases, NNs struggle with slower convergence rates, leading to longer recovery times when adapting to new distributions. The critical challenge now lies in integrating these components into autonomous, resource-limited learning systems, balancing the stability-plasticity and forward-backward transfer trade-offs. Additionally, complementary approaches not covered in this survey, such as federated learning, offer promising directions for scaling model training across multiple edge devices while preserving data privacy and handling heterogeneous distributions.

\bibliographystyle{cas-model2-names}

\bibliography{cas-refs}

\begin{thebibliography}{165}
\expandafter\ifx\csname natexlab\endcsname\relax\def\natexlab#1{#1}\fi
\providecommand{\url}[1]{\texttt{#1}}
\providecommand{\href}[2]{#2}
\providecommand{\path}[1]{#1}
\providecommand{\DOIprefix}{doi:}
\providecommand{\ArXivprefix}{arXiv:}
\providecommand{\URLprefix}{URL: }
\providecommand{\Pubmedprefix}{pmid:}
\providecommand{\doi}[1]{\href{http://dx.doi.org/#1}{\path{#1}}}
\providecommand{\Pubmed}[1]{\href{pmid:#1}{\path{#1}}}
\providecommand{\bibinfo}[2]{#2}
\ifx\xfnm\relax \def\xfnm[#1]{\unskip,\space#1}\fi
%Type = Inproceedings
\bibitem[{Abati et~al.(2020)Abati, Tomczak, Blankevoort, Calderara, Cucchiara and Bejnordi}]{abati2020conditional}
\bibinfo{author}{Abati, D.}, \bibinfo{author}{Tomczak, J.}, \bibinfo{author}{Blankevoort, T.}, \bibinfo{author}{Calderara, S.}, \bibinfo{author}{Cucchiara, R.}, \bibinfo{author}{Bejnordi, B.E.}, \bibinfo{year}{2020}.
\newblock \bibinfo{title}{Conditional channel gated networks for task-aware continual learning}, in: \bibinfo{booktitle}{Proceedings of the IEEE/CVF conference on computer vision and pattern recognition}, pp. \bibinfo{pages}{3931--3940}.
%Type = Article
\bibitem[{Achille et~al.(2018)Achille, Rovere and Santoro}]{achille2018critical}
\bibinfo{author}{Achille, A.}, \bibinfo{author}{Rovere, M.}, \bibinfo{author}{Santoro, S.}, \bibinfo{year}{2018}.
\newblock \bibinfo{title}{Critical learning periods in deep networks}.
\newblock \bibinfo{journal}{arXiv preprint arXiv:1711.08856} .
%Type = Inproceedings
\bibitem[{Aljundi et~al.(2019)Aljundi, Belilovsky, Tuytelaars, Charlin, Caccia, Lin and Page-Caccia}]{aljundi2019online}
\bibinfo{author}{Aljundi, R.}, \bibinfo{author}{Belilovsky, E.}, \bibinfo{author}{Tuytelaars, T.}, \bibinfo{author}{Charlin, L.}, \bibinfo{author}{Caccia, M.}, \bibinfo{author}{Lin, M.}, \bibinfo{author}{Page-Caccia, L.}, \bibinfo{year}{2019}.
\newblock \bibinfo{title}{Online continual learning with maximal interfered retrieval}, in: \bibinfo{booktitle}{Advances in Neural Information Processing Systems}, pp. \bibinfo{pages}{11849--11860}.
%Type = Article
\bibitem[{Almeida et~al.(2018)Almeida, Oliveira, Britto~Jr and Sabourin}]{almeida2018adapting}
\bibinfo{author}{Almeida, P.R.}, \bibinfo{author}{Oliveira, L.S.}, \bibinfo{author}{Britto~Jr, A.S.}, \bibinfo{author}{Sabourin, R.}, \bibinfo{year}{2018}.
\newblock \bibinfo{title}{Adapting dynamic classifier selection for concept drift}.
\newblock \bibinfo{journal}{Expert Systems with Applications} \bibinfo{volume}{104}, \bibinfo{pages}{67--85}.
%Type = Inproceedings
\bibitem[{Anderson et~al.(2016)Anderson, Koh and Dobbie}]{anderson2016recurring}
\bibinfo{author}{Anderson, R.}, \bibinfo{author}{Koh, Y.S.}, \bibinfo{author}{Dobbie, G.}, \bibinfo{year}{2016}.
\newblock \bibinfo{title}{Recurring concept meta-learning for evolving data streams}, in: \bibinfo{booktitle}{Pacific-Asia Conference on Knowledge Discovery and Data Mining}, \bibinfo{organization}{Springer}. pp. \bibinfo{pages}{3--14}.
%Type = Article
\bibitem[{Ankorion(2005)}]{ankorion2005change}
\bibinfo{author}{Ankorion, I.}, \bibinfo{year}{2005}.
\newblock \bibinfo{title}{Change data capture: Efficient etl for real-time bi}.
\newblock \bibinfo{journal}{Information Management} \bibinfo{volume}{15}.
%Type = Inproceedings
\bibitem[{Arik and Pfister(2021)}]{tabnet2019arik}
\bibinfo{author}{Arik, S.O.}, \bibinfo{author}{Pfister, T.}, \bibinfo{year}{2021}.
\newblock \bibinfo{title}{Tabnet: Attentive interpretable tabular learning}, in: \bibinfo{booktitle}{Proceedings of the AAAI Conference on Artificial Intelligence}, pp. \bibinfo{pages}{6679--6687}.
%Type = Inproceedings
\bibitem[{Ash and Adams(2020)}]{ash2020warm}
\bibinfo{author}{Ash, J.}, \bibinfo{author}{Adams, R.P.}, \bibinfo{year}{2020}.
\newblock \bibinfo{title}{On warm-starting neural network training}, in: \bibinfo{booktitle}{Advances in Neural Information Processing Systems}, pp. \bibinfo{pages}{3884--3894}.
%Type = Article
\bibitem[{Badirli et~al.(2020)Badirli, Liu, Xing, Bhowmik, Doan and Keerthi}]{badirli2020gradient}
\bibinfo{author}{Badirli, S.}, \bibinfo{author}{Liu, X.}, \bibinfo{author}{Xing, Z.}, \bibinfo{author}{Bhowmik, A.}, \bibinfo{author}{Doan, K.}, \bibinfo{author}{Keerthi, S.S.}, \bibinfo{year}{2020}.
\newblock \bibinfo{title}{Gradient boosting neural networks: Grownet}.
\newblock \bibinfo{journal}{arXiv preprint arXiv:2002.07971} .
%Type = Inproceedings
\bibitem[{Bang et~al.(2021)Bang, Kim, Yoo, Ha and Choi}]{bang2021rainbow}
\bibinfo{author}{Bang, J.}, \bibinfo{author}{Kim, H.}, \bibinfo{author}{Yoo, Y.}, \bibinfo{author}{Ha, J.W.}, \bibinfo{author}{Choi, J.}, \bibinfo{year}{2021}.
\newblock \bibinfo{title}{Rainbow memory: Continual learning with a memory of diverse samples}, in: \bibinfo{booktitle}{Proceedings of the IEEE/CVF Conference on Computer Vision and Pattern Recognition}, pp. \bibinfo{pages}{8218--8227}.
%Type = Article
\bibitem[{Barddal and Enembreck(2020)}]{barddal2020regularized}
\bibinfo{author}{Barddal, J.P.}, \bibinfo{author}{Enembreck, F.}, \bibinfo{year}{2020}.
\newblock \bibinfo{title}{Regularized and incremental decision trees for data streams}.
\newblock \bibinfo{journal}{Annals of Telecommunications} \bibinfo{volume}{75}, \bibinfo{pages}{493--503}.
%Type = Inproceedings
\bibitem[{Bifet and Gavalda(2009)}]{bifet2009adaptive}
\bibinfo{author}{Bifet, A.}, \bibinfo{author}{Gavalda, R.}, \bibinfo{year}{2009}.
\newblock \bibinfo{title}{Adaptive learning from evolving data streams}, in: \bibinfo{booktitle}{Advances in Intelligent Data Analysis VIII: 8th International Symposium on Intelligent Data Analysis, IDA 2009, Lyon, France, August 31-September 2, 2009. Proceedings 8}, \bibinfo{organization}{Springer}. pp. \bibinfo{pages}{249--260}.
%Type = Inproceedings
\bibitem[{Bonet et~al.(2024)Bonet, Montserrat, Gir{\'o}-i Nieto and Ioannidis}]{bonet2024hyperfast}
\bibinfo{author}{Bonet, D.}, \bibinfo{author}{Montserrat, D.M.}, \bibinfo{author}{Gir{\'o}-i Nieto, X.}, \bibinfo{author}{Ioannidis, A.G.}, \bibinfo{year}{2024}.
\newblock \bibinfo{title}{Hyperfast: Instant classification for tabular data}, in: \bibinfo{booktitle}{Proceedings of the AAAI Conference on Artificial Intelligence}, pp. \bibinfo{pages}{11114--11123}.
%Type = Article
\bibitem[{Borisov et~al.(2022)Borisov, Leemann, Se{\ss}ler, Haug, Pawelczyk and Kasneci}]{borisov2022deep}
\bibinfo{author}{Borisov, V.}, \bibinfo{author}{Leemann, T.}, \bibinfo{author}{Se{\ss}ler, K.}, \bibinfo{author}{Haug, J.}, \bibinfo{author}{Pawelczyk, M.}, \bibinfo{author}{Kasneci, G.}, \bibinfo{year}{2022}.
\newblock \bibinfo{title}{Deep neural networks and tabular data: A survey}.
\newblock \bibinfo{journal}{IEEE transactions on neural networks and learning systems} .
%Type = Article
\bibitem[{Buzzega et~al.(2020)Buzzega, Boschini, Porrello, Abati and Calderara}]{buzzega2020dark}
\bibinfo{author}{Buzzega, P.}, \bibinfo{author}{Boschini, M.}, \bibinfo{author}{Porrello, A.}, \bibinfo{author}{Abati, D.}, \bibinfo{author}{Calderara, S.}, \bibinfo{year}{2020}.
\newblock \bibinfo{title}{Dark experience for general continual learning: a strong, simple baseline}.
\newblock \bibinfo{journal}{Advances in neural information processing systems} \bibinfo{volume}{33}, \bibinfo{pages}{15920--15930}.
%Type = Article
\bibitem[{Caccia et~al.(2022)Caccia, Belilovsky, Caccia and Pineau}]{caccia2022new}
\bibinfo{author}{Caccia, M.}, \bibinfo{author}{Belilovsky, E.}, \bibinfo{author}{Caccia, M.}, \bibinfo{author}{Pineau, J.}, \bibinfo{year}{2022}.
\newblock \bibinfo{title}{New insights on reducing abrupt representation change in online continual learning}.
\newblock \bibinfo{journal}{arXiv preprint arXiv:2203.06389} .
%Type = Article
\bibitem[{Cha et~al.(2022)Cha, Shim, Kim, Lee, Lee and Moon}]{cha2022towards}
\bibinfo{author}{Cha, S.}, \bibinfo{author}{Shim, D.}, \bibinfo{author}{Kim, H.}, \bibinfo{author}{Lee, M.}, \bibinfo{author}{Lee, H.}, \bibinfo{author}{Moon, T.}, \bibinfo{year}{2022}.
\newblock \bibinfo{title}{Towards more objective evaluation of class incremental learning: Representation learning perspective}.
\newblock \bibinfo{journal}{arXiv preprint arXiv:2206.08101} .
%Type = Inproceedings
\bibitem[{Chaudhry et~al.(2018)Chaudhry, Ranzato, Rohrbach and Elhoseiny}]{chaudhry2018efficient}
\bibinfo{author}{Chaudhry, A.}, \bibinfo{author}{Ranzato, M.}, \bibinfo{author}{Rohrbach, M.}, \bibinfo{author}{Elhoseiny, M.}, \bibinfo{year}{2018}.
\newblock \bibinfo{title}{Efficient lifelong learning with a-gem}, in: \bibinfo{booktitle}{International Conference on Learning Representations}.
%Type = Article
\bibitem[{Chaudhry et~al.(2019)Chaudhry, Rohrbach, Elhoseiny, Ajanthan, Dokania, Torr and Ranzato}]{chaudhry2019tiny}
\bibinfo{author}{Chaudhry, A.}, \bibinfo{author}{Rohrbach, M.}, \bibinfo{author}{Elhoseiny, M.}, \bibinfo{author}{Ajanthan, T.}, \bibinfo{author}{Dokania, P.K.}, \bibinfo{author}{Torr, P.H.}, \bibinfo{author}{Ranzato, M.}, \bibinfo{year}{2019}.
\newblock \bibinfo{title}{On tiny episodic memories in continual learning}.
\newblock \bibinfo{journal}{arXiv preprint arXiv:1902.10486} .
%Type = Article
\bibitem[{Cheung et~al.(2019)Cheung, Terekhov, Chen, Agrawal and Olshausen}]{cheung2019superposition}
\bibinfo{author}{Cheung, B.}, \bibinfo{author}{Terekhov, A.}, \bibinfo{author}{Chen, Y.}, \bibinfo{author}{Agrawal, P.}, \bibinfo{author}{Olshausen, B.}, \bibinfo{year}{2019}.
\newblock \bibinfo{title}{Superposition of many models into one}.
\newblock \bibinfo{journal}{Advances in Neural Information Processing Systems} \bibinfo{volume}{32}.
%Type = Article
\bibitem[{da~Costa et~al.(2018)da~Costa, de~Leon~Ferreira, Junior et~al.}]{da2018strict}
\bibinfo{author}{da~Costa, V.G.T.}, \bibinfo{author}{de~Leon~Ferreira, A.C.P.}, \bibinfo{author}{Junior, S.B.}, et~al., \bibinfo{year}{2018}.
\newblock \bibinfo{title}{Strict very fast decision tree: a memory conservative algorithm for data stream mining}.
\newblock \bibinfo{journal}{Pattern Recognition Letters} \bibinfo{volume}{116}, \bibinfo{pages}{22--28}.
%Type = Inproceedings
\bibitem[{Courbariaux et~al.(2015)Courbariaux, Bengio and David}]{courbariaux2015binaryconnect}
\bibinfo{author}{Courbariaux, M.}, \bibinfo{author}{Bengio, Y.}, \bibinfo{author}{David, J.P.}, \bibinfo{year}{2015}.
\newblock \bibinfo{title}{Binaryconnect: Training deep neural networks with binary weights during propagations}, in: \bibinfo{booktitle}{Advances in neural information processing systems}, pp. \bibinfo{pages}{3123--3131}.
%Type = Article
\bibitem[{De~Lange and Tuytelaars(2021)}]{de2021continual}
\bibinfo{author}{De~Lange, M.}, \bibinfo{author}{Tuytelaars, T.}, \bibinfo{year}{2021}.
\newblock \bibinfo{title}{Continual prototype evolution: Learning online from non-stationary data streams}.
\newblock \bibinfo{journal}{Proceedings of the IEEE/CVF International Conference on Computer Vision} , \bibinfo{pages}{8250--8259}.
%Type = Article
\bibitem[{De~Rosa and Cesa-Bianchi(2017)}]{de2017confidence}
\bibinfo{author}{De~Rosa, R.}, \bibinfo{author}{Cesa-Bianchi, N.}, \bibinfo{year}{2017}.
\newblock \bibinfo{title}{Confidence decision trees via online and active learning for streaming data}.
\newblock \bibinfo{journal}{Journal of Artificial Intelligence Research} \bibinfo{volume}{60}, \bibinfo{pages}{1031--1055}.
%Type = Inproceedings
\bibitem[{Debnath et~al.(2008)Debnath, Lilja and Mokbel}]{debnath2008sard}
\bibinfo{author}{Debnath, B.K.}, \bibinfo{author}{Lilja, D.J.}, \bibinfo{author}{Mokbel, M.F.}, \bibinfo{year}{2008}.
\newblock \bibinfo{title}{{SARD: A statistical approach for ranking database tuning parameters}}, in: \bibinfo{booktitle}{2008 IEEE 24th International Conference on Data Engineering Workshop}, \bibinfo{organization}{IEEE}. \bibinfo{publisher}{IEEE}. pp. \bibinfo{pages}{11--18}.
%Type = Inproceedings
\bibitem[{Denton et~al.(2014)Denton, Zaremba, Bruna, LeCun and Fergus}]{denton2014exploiting}
\bibinfo{author}{Denton, E.L.}, \bibinfo{author}{Zaremba, W.}, \bibinfo{author}{Bruna, J.}, \bibinfo{author}{LeCun, Y.}, \bibinfo{author}{Fergus, R.}, \bibinfo{year}{2014}.
\newblock \bibinfo{title}{Exploiting linear structure within convolutional networks for efficient evaluation}, in: \bibinfo{booktitle}{Advances in neural information processing systems}.
%Type = Inproceedings
\bibitem[{Dohare et~al.(2021)Dohare, Hernandez-Garcia, Lacoste and Weiss}]{dohare2021continual}
\bibinfo{author}{Dohare, S.}, \bibinfo{author}{Hernandez-Garcia, A.}, \bibinfo{author}{Lacoste, A.}, \bibinfo{author}{Weiss, M.}, \bibinfo{year}{2021}.
\newblock \bibinfo{title}{Continual backprop: Stochastic gradient descent with persistent randomness}, in: \bibinfo{booktitle}{International Conference on Machine Learning}, pp. \bibinfo{pages}{2660--2670}.
%Type = Inproceedings
\bibitem[{Domingos and Hulten(2000)}]{domingos2000mining}
\bibinfo{author}{Domingos, P.}, \bibinfo{author}{Hulten, G.}, \bibinfo{year}{2000}.
\newblock \bibinfo{title}{Mining high-speed data streams}, in: \bibinfo{booktitle}{Proceedings of the sixth ACM SIGKDD international conference on Knowledge discovery and data mining}, pp. \bibinfo{pages}{71--80}.
%Type = Inproceedings
\bibitem[{Domingos and Hulten(2001)}]{domingos2001catching}
\bibinfo{author}{Domingos, P.M.}, \bibinfo{author}{Hulten, G.}, \bibinfo{year}{2001}.
\newblock \bibinfo{title}{Catching up with the data: Research issues in mining data streams}, in: \bibinfo{booktitle}{2001 ACM SIGMOD Workshop on Research Issues in Data Mining and Knowledge Discovery}, \bibinfo{address}{Santa Barbara, CA, USA}.
%Type = Inproceedings
\bibitem[{Dong et~al.(2017)Dong, Chen and Pan}]{dong2017learning}
\bibinfo{author}{Dong, X.}, \bibinfo{author}{Chen, S.}, \bibinfo{author}{Pan, S.}, \bibinfo{year}{2017}.
\newblock \bibinfo{title}{Learning to prune deep neural networks via layer-wise optimal brain surgeon}, in: \bibinfo{booktitle}{Advances in Neural Information Processing Systems}, pp. \bibinfo{pages}{4857--4867}.
%Type = Article
\bibitem[{Ducange et~al.(2021)Ducange, Marcelloni, Pecori et~al.}]{ducange2021fuzzy}
\bibinfo{author}{Ducange, P.}, \bibinfo{author}{Marcelloni, F.}, \bibinfo{author}{Pecori, R.}, et~al., \bibinfo{year}{2021}.
\newblock \bibinfo{title}{Fuzzy hoeffding decision tree for data stream classification.}
\newblock \bibinfo{journal}{Int. J. Comput. Intell. Syst.} \bibinfo{volume}{14}, \bibinfo{pages}{946--964}.
%Type = Inproceedings
\bibitem[{Duda et~al.(2014)Duda, Jaworski, Pietruczuk and Rutkowski}]{duda2014novel}
\bibinfo{author}{Duda, P.}, \bibinfo{author}{Jaworski, M.}, \bibinfo{author}{Pietruczuk, L.}, \bibinfo{author}{Rutkowski, L.}, \bibinfo{year}{2014}.
\newblock \bibinfo{title}{A novel application of hoeffding's inequality to decision trees construction for data streams}, in: \bibinfo{booktitle}{2014 International Joint Conference on Neural Networks (IJCNN)}, \bibinfo{organization}{IEEE}. pp. \bibinfo{pages}{3324--3330}.
%Type = Article
\bibitem[{Faria et~al.(2016)Faria, Gama and Carvalho}]{faria2016minas}
\bibinfo{author}{Faria, E.R.d.}, \bibinfo{author}{Gama, J.}, \bibinfo{author}{Carvalho, A.C.P.d.L.F.}, \bibinfo{year}{2016}.
\newblock \bibinfo{title}{Minas: multiclass learning algorithm for novelty detection in data streams}.
\newblock \bibinfo{journal}{Data Mining and Knowledge Discovery} \bibinfo{volume}{30}, \bibinfo{pages}{640--680}.
%Type = Article
\bibitem[{Frias-Blanco et~al.(2014)Frias-Blanco, del Campo-{\'A}vila, Ramos-Jimenez, Morales-Bueno, Ortiz-Diaz and Caballero-Mota}]{frias2014online}
\bibinfo{author}{Frias-Blanco, I.}, \bibinfo{author}{del Campo-{\'A}vila, J.}, \bibinfo{author}{Ramos-Jimenez, G.}, \bibinfo{author}{Morales-Bueno, R.}, \bibinfo{author}{Ortiz-Diaz, A.}, \bibinfo{author}{Caballero-Mota, Y.}, \bibinfo{year}{2014}.
\newblock \bibinfo{title}{Online and non-parametric drift detection methods based on hoeffding’s bounds}.
\newblock \bibinfo{journal}{IEEE Transactions on Knowledge and Data Engineering} \bibinfo{volume}{27}, \bibinfo{pages}{810--823}.
%Type = Article
\bibitem[{Frosst and Hinton(2017)}]{frosst2017distilling}
\bibinfo{author}{Frosst, N.}, \bibinfo{author}{Hinton, G.}, \bibinfo{year}{2017}.
\newblock \bibinfo{title}{Distilling a neural network into a soft decision tree}.
\newblock \bibinfo{journal}{arXiv preprint arXiv:1711.09784} .
%Type = Inproceedings
\bibitem[{Gama et~al.(2004)Gama, Medas and Rocha}]{gama2004forest}
\bibinfo{author}{Gama, J.}, \bibinfo{author}{Medas, P.}, \bibinfo{author}{Rocha, R.}, \bibinfo{year}{2004}.
\newblock \bibinfo{title}{Forest trees for on-line data}, in: \bibinfo{booktitle}{Proceedings of the 2004 ACM symposium on Applied computing}, pp. \bibinfo{pages}{632--636}.
%Type = Inproceedings
\bibitem[{Gama et~al.(2005)Gama, Medas and Rodrigues}]{gama2005learning}
\bibinfo{author}{Gama, J.}, \bibinfo{author}{Medas, P.}, \bibinfo{author}{Rodrigues, P.}, \bibinfo{year}{2005}.
\newblock \bibinfo{title}{Learning decision trees from dynamic data streams}, in: \bibinfo{booktitle}{Proceedings of the 2005 ACM Symposium on Applied computing}, pp. \bibinfo{pages}{573--577}.
%Type = Inproceedings
\bibitem[{Gama and Pinto(2006)}]{gama2006discretization}
\bibinfo{author}{Gama, J.}, \bibinfo{author}{Pinto, C.}, \bibinfo{year}{2006}.
\newblock \bibinfo{title}{Discretization from data streams: applications to histograms and data mining}, in: \bibinfo{booktitle}{Proceedings of the 2006 ACM symposium on Applied computing}, pp. \bibinfo{pages}{662--667}.
%Type = Inproceedings
\bibitem[{Gama et~al.(2003)Gama, Rocha and Medas}]{gama2003accurate}
\bibinfo{author}{Gama, J.}, \bibinfo{author}{Rocha, R.}, \bibinfo{author}{Medas, P.}, \bibinfo{year}{2003}.
\newblock \bibinfo{title}{Accurate decision trees for mining high-speed data streams}, in: \bibinfo{booktitle}{Proceedings of the ninth ACM SIGKDD international conference on Knowledge discovery and data mining}, pp. \bibinfo{pages}{523--528}.
%Type = Article
\bibitem[{Gama et~al.(2013)Gama, Sebastiao and Rodrigues}]{gama2013evaluating}
\bibinfo{author}{Gama, J.}, \bibinfo{author}{Sebastiao, R.}, \bibinfo{author}{Rodrigues, P.P.}, \bibinfo{year}{2013}.
\newblock \bibinfo{title}{On evaluating stream learning algorithms}.
\newblock \bibinfo{journal}{Machine learning} \bibinfo{volume}{90}, \bibinfo{pages}{317--346}.
%Type = Article
\bibitem[{Garcia-Martin et~al.(2022)Garcia-Martin, Bifet, Lavesson, K{\"o}nig and Linusson}]{garcia2022green}
\bibinfo{author}{Garcia-Martin, E.}, \bibinfo{author}{Bifet, A.}, \bibinfo{author}{Lavesson, N.}, \bibinfo{author}{K{\"o}nig, R.}, \bibinfo{author}{Linusson, H.}, \bibinfo{year}{2022}.
\newblock \bibinfo{title}{Green accelerated hoeffding tree}.
\newblock \bibinfo{journal}{arXiv preprint arXiv:2205.03184} .
%Type = Inproceedings
\bibitem[{Garc{\'\i}a-Mart{\'\i}n et~al.(2018)Garc{\'\i}a-Mart{\'\i}n, Lavesson, Grahn, Casalicchio and Boeva}]{garcia2018hoeffding}
\bibinfo{author}{Garc{\'\i}a-Mart{\'\i}n, E.}, \bibinfo{author}{Lavesson, N.}, \bibinfo{author}{Grahn, H.}, \bibinfo{author}{Casalicchio, E.}, \bibinfo{author}{Boeva, V.}, \bibinfo{year}{2018}.
\newblock \bibinfo{title}{Hoeffding trees with nmin adaptation}, in: \bibinfo{booktitle}{2018 IEEE 5th International Conference on Data Science and Advanced Analytics (DSAA)}, \bibinfo{organization}{IEEE}. pp. \bibinfo{pages}{70--79}.
%Type = Article
\bibitem[{Geng et~al.(2020)Geng, Huang and Chen}]{geng2020recent}
\bibinfo{author}{Geng, C.}, \bibinfo{author}{Huang, S.j.}, \bibinfo{author}{Chen, S.}, \bibinfo{year}{2020}.
\newblock \bibinfo{title}{Recent advances in open set recognition: A survey}.
\newblock \bibinfo{journal}{IEEE transactions on pattern analysis and machine intelligence} \bibinfo{volume}{43}, \bibinfo{pages}{3614--3631}.
%Type = Article
\bibitem[{Gomes et~al.(2017)Gomes, Bifet, Read, Barddal, Enembreck, Pfharinger, Holmes and Abdessalem}]{gomes2017adaptive}
\bibinfo{author}{Gomes, H.M.}, \bibinfo{author}{Bifet, A.}, \bibinfo{author}{Read, J.}, \bibinfo{author}{Barddal, J.P.}, \bibinfo{author}{Enembreck, F.}, \bibinfo{author}{Pfharinger, B.}, \bibinfo{author}{Holmes, G.}, \bibinfo{author}{Abdessalem, T.}, \bibinfo{year}{2017}.
\newblock \bibinfo{title}{Adaptive random forests for evolving data stream classification}.
\newblock \bibinfo{journal}{Machine Learning} \bibinfo{volume}{106}, \bibinfo{pages}{1469--1495}.
%Type = Article
\bibitem[{Gorishniy et~al.(2022)Gorishniy, Rubachev and Babenko}]{gorishniy2022embeddings}
\bibinfo{author}{Gorishniy, Y.}, \bibinfo{author}{Rubachev, I.}, \bibinfo{author}{Babenko, A.}, \bibinfo{year}{2022}.
\newblock \bibinfo{title}{On embeddings for numerical features in tabular deep learning}.
\newblock \bibinfo{journal}{Advances in Neural Information Processing Systems} \bibinfo{volume}{35}, \bibinfo{pages}{24991--25004}.
%Type = Article
\bibitem[{G{\"o}z{\"u}a{\c{c}}{\i}k and Can(2021)}]{gozuaccik2021concept}
\bibinfo{author}{G{\"o}z{\"u}a{\c{c}}{\i}k, {\"O}.}, \bibinfo{author}{Can, F.}, \bibinfo{year}{2021}.
\newblock \bibinfo{title}{Concept learning using one-class classifiers for implicit drift detection in evolving data streams}.
\newblock \bibinfo{journal}{Artificial Intelligence Review} \bibinfo{volume}{54}, \bibinfo{pages}{3725--3747}.
%Type = Article
\bibitem[{Grinsztajn et~al.(2022)Grinsztajn, Oyallon and Varoquaux}]{grinsztajn2022tree}
\bibinfo{author}{Grinsztajn, L.}, \bibinfo{author}{Oyallon, E.}, \bibinfo{author}{Varoquaux, G.}, \bibinfo{year}{2022}.
\newblock \bibinfo{title}{Why do tree-based models still outperform deep learning on tabular data?}
\newblock \bibinfo{journal}{arXiv preprint arXiv:2207.08815} .
%Type = Article
\bibitem[{Gunasekara et~al.(2024)Gunasekara, Pfahringer, Gomes and Bifet}]{gunasekara2024gradient}
\bibinfo{author}{Gunasekara, N.}, \bibinfo{author}{Pfahringer, B.}, \bibinfo{author}{Gomes, H.}, \bibinfo{author}{Bifet, A.}, \bibinfo{year}{2024}.
\newblock \bibinfo{title}{Gradient boosted trees for evolving data streams}.
\newblock \bibinfo{journal}{Machine Learning} \bibinfo{volume}{113}, \bibinfo{pages}{3325--3352}.
%Type = Inproceedings
\bibitem[{Guo et~al.(2023)Guo, Liu and Zhao}]{guo2023dealing}
\bibinfo{author}{Guo, Y.}, \bibinfo{author}{Liu, B.}, \bibinfo{author}{Zhao, D.}, \bibinfo{year}{2023}.
\newblock \bibinfo{title}{Dealing with cross-task class discrimination in online continual learning}, in: \bibinfo{booktitle}{Proceedings of the IEEE/CVF Conference on Computer Vision and Pattern Recognition}, pp. \bibinfo{pages}{20446--20455}.
%Type = Article
\bibitem[{Han et~al.(2022)Han, Hu, Huang, Jiang and Zhao}]{han2022adbench}
\bibinfo{author}{Han, S.}, \bibinfo{author}{Hu, X.}, \bibinfo{author}{Huang, H.}, \bibinfo{author}{Jiang, M.}, \bibinfo{author}{Zhao, Y.}, \bibinfo{year}{2022}.
\newblock \bibinfo{title}{Adbench: Anomaly detection benchmark}.
\newblock \bibinfo{journal}{Advances in Neural Information Processing Systems} \bibinfo{volume}{35}, \bibinfo{pages}{32142--32159}.
%Type = Inproceedings
\bibitem[{Han et~al.(2015)Han, Pool, Tran and Dally}]{han2015learning}
\bibinfo{author}{Han, S.}, \bibinfo{author}{Pool, J.}, \bibinfo{author}{Tran, J.}, \bibinfo{author}{Dally, W.}, \bibinfo{year}{2015}.
\newblock \bibinfo{title}{Learning both weights and connections for efficient neural network}, in: \bibinfo{booktitle}{Advances in neural information processing systems}, pp. \bibinfo{pages}{1135--1143}.
%Type = Inproceedings
\bibitem[{Haque et~al.(2016)Haque, Khan and Baron}]{haque2016sand}
\bibinfo{author}{Haque, A.}, \bibinfo{author}{Khan, L.}, \bibinfo{author}{Baron, M.}, \bibinfo{year}{2016}.
\newblock \bibinfo{title}{Sand: Semi-supervised adaptive novel class detection and classification over data stream}, in: \bibinfo{booktitle}{Proceedings of the AAAI Conference on Artificial Intelligence}.
%Type = Article
\bibitem[{Harrison et~al.(2020)Harrison, Sharma, Finn and Pavone}]{harrison2019continuous}
\bibinfo{author}{Harrison, J.}, \bibinfo{author}{Sharma, A.}, \bibinfo{author}{Finn, C.}, \bibinfo{author}{Pavone, M.}, \bibinfo{year}{2020}.
\newblock \bibinfo{title}{Continuous meta-learning without tasks}.
\newblock \bibinfo{journal}{Advances in neural information processing systems} \bibinfo{volume}{33}, \bibinfo{pages}{17571--17581}.
%Type = Article
\bibitem[{Hayes et~al.(2020)Hayes, Kafle, Shrestha, Acharya and Kanan}]{hayes2021remind}
\bibinfo{author}{Hayes, T.L.}, \bibinfo{author}{Kafle, K.}, \bibinfo{author}{Shrestha, R.}, \bibinfo{author}{Acharya, M.}, \bibinfo{author}{Kanan, C.}, \bibinfo{year}{2020}.
\newblock \bibinfo{title}{Remind your neural network to prevent catastrophic forgetting}.
\newblock \bibinfo{journal}{European Conference on Computer Vision} , \bibinfo{pages}{466--483}.
%Type = Article
\bibitem[{Hess et~al.(2023)Hess, Verwimp, van~de Ven and Tuytelaars}]{hess2023knowledge}
\bibinfo{author}{Hess, T.}, \bibinfo{author}{Verwimp, E.}, \bibinfo{author}{van~de Ven, G.M.}, \bibinfo{author}{Tuytelaars, T.}, \bibinfo{year}{2023}.
\newblock \bibinfo{title}{Knowledge accumulation in continually learned representations and the issue of feature forgetting}.
\newblock \bibinfo{journal}{arXiv preprint arXiv:2304.00933} .
%Type = Inproceedings
\bibitem[{Heyden et~al.(2024)Heyden, Gomes, Fouch{\'e}, Pfahringer and B{\"o}hm}]{heyden2024leveraging}
\bibinfo{author}{Heyden, M.}, \bibinfo{author}{Gomes, H.M.}, \bibinfo{author}{Fouch{\'e}, E.}, \bibinfo{author}{Pfahringer, B.}, \bibinfo{author}{B{\"o}hm, K.}, \bibinfo{year}{2024}.
\newblock \bibinfo{title}{Leveraging plasticity in incremental decision trees}, in: \bibinfo{booktitle}{Joint European Conference on Machine Learning and Knowledge Discovery in Databases}, \bibinfo{organization}{Springer}. pp. \bibinfo{pages}{38--54}.
%Type = Article
\bibitem[{Hollmann et~al.(2025)Hollmann, M{\"u}ller, Purucker, Krishnakumar, K{\"o}rfer, Hoo, Schirrmeister and Hutter}]{hollmann2025accurate}
\bibinfo{author}{Hollmann, N.}, \bibinfo{author}{M{\"u}ller, S.}, \bibinfo{author}{Purucker, L.}, \bibinfo{author}{Krishnakumar, A.}, \bibinfo{author}{K{\"o}rfer, M.}, \bibinfo{author}{Hoo, S.B.}, \bibinfo{author}{Schirrmeister, R.T.}, \bibinfo{author}{Hutter, F.}, \bibinfo{year}{2025}.
\newblock \bibinfo{title}{Accurate predictions on small data with a tabular foundation model}.
\newblock \bibinfo{journal}{Nature} \bibinfo{volume}{637}, \bibinfo{pages}{319--326}.
%Type = Inproceedings
\bibitem[{Holmes et~al.(2005)Holmes, Richard and Pfahringer}]{holmes2005tie}
\bibinfo{author}{Holmes, G.}, \bibinfo{author}{Richard, K.}, \bibinfo{author}{Pfahringer, B.}, \bibinfo{year}{2005}.
\newblock \bibinfo{title}{Tie-breaking in hoeffding trees}, \bibinfo{organization}{ECML/PKDD}.
%Type = Article
\bibitem[{Howard(2017)}]{howard2017mobilenets}
\bibinfo{author}{Howard, A.G.}, \bibinfo{year}{2017}.
\newblock \bibinfo{title}{Mobilenets: Efficient convolutional neural networks for mobile vision applications}.
\newblock \bibinfo{journal}{arXiv preprint arXiv:1704.04861} .
%Type = Article
\bibitem[{Huang et~al.(2020)Huang, Khetan, Cvitkovic and Karnin}]{huang2020tabtransformer}
\bibinfo{author}{Huang, X.}, \bibinfo{author}{Khetan, A.}, \bibinfo{author}{Cvitkovic, M.}, \bibinfo{author}{Karnin, Z.}, \bibinfo{year}{2020}.
\newblock \bibinfo{title}{Tabtransformer: Tabular data modeling using contextual embeddings}.
\newblock \bibinfo{journal}{arXiv preprint arXiv:2012.06678} .
%Type = Inproceedings
\bibitem[{Hulten et~al.(2001)Hulten, Spencer and Domingos}]{hulten2001mining}
\bibinfo{author}{Hulten, G.}, \bibinfo{author}{Spencer, L.}, \bibinfo{author}{Domingos, P.}, \bibinfo{year}{2001}.
\newblock \bibinfo{title}{Mining time-changing data streams}, in: \bibinfo{booktitle}{Proceedings of the seventh ACM SIGKDD international conference on Knowledge discovery and data mining}, pp. \bibinfo{pages}{97--106}.
%Type = Article
\bibitem[{Iandola(2016)}]{iandola2016squeezenet}
\bibinfo{author}{Iandola, F.N.}, \bibinfo{year}{2016}.
\newblock \bibinfo{title}{Squeezenet: Alexnet-level accuracy with 50x fewer parameters and< 0.5 mb model size}.
\newblock \bibinfo{journal}{arXiv preprint arXiv:1602.07360} .
%Type = Article
\bibitem[{Ikonomovska et~al.(2011)Ikonomovska, Gama and D{\v{z}}eroski}]{ikonomovska2011learning}
\bibinfo{author}{Ikonomovska, E.}, \bibinfo{author}{Gama, J.}, \bibinfo{author}{D{\v{z}}eroski, S.}, \bibinfo{year}{2011}.
\newblock \bibinfo{title}{Learning model trees from evolving data streams}.
\newblock \bibinfo{journal}{Data mining and knowledge discovery} \bibinfo{volume}{23}, \bibinfo{pages}{128--168}.
%Type = Article
\bibitem[{Jaderberg et~al.(2014)Jaderberg, Vedaldi and Zisserman}]{jaderberg2014speeding}
\bibinfo{author}{Jaderberg, M.}, \bibinfo{author}{Vedaldi, A.}, \bibinfo{author}{Zisserman, A.}, \bibinfo{year}{2014}.
\newblock \bibinfo{title}{Speeding up convolutional neural networks with low rank expansions}.
\newblock \bibinfo{journal}{arXiv preprint arXiv:1405.3866} .
%Type = Article
\bibitem[{Jeffares et~al.(2023)Jeffares, Liu, Crabb{\'e}, Imrie and van~der Schaar}]{jeffares2023tangos}
\bibinfo{author}{Jeffares, A.}, \bibinfo{author}{Liu, T.}, \bibinfo{author}{Crabb{\'e}, J.}, \bibinfo{author}{Imrie, F.}, \bibinfo{author}{van~der Schaar, M.}, \bibinfo{year}{2023}.
\newblock \bibinfo{title}{Tangos: Regularizing tabular neural networks through gradient orthogonalization and specialization}.
\newblock \bibinfo{journal}{arXiv preprint arXiv:2303.05506} .
%Type = Article
\bibitem[{Katakis et~al.(2010)Katakis, Tsoumakas and Vlahavas}]{katakis2010tracking}
\bibinfo{author}{Katakis, I.}, \bibinfo{author}{Tsoumakas, G.}, \bibinfo{author}{Vlahavas, I.}, \bibinfo{year}{2010}.
\newblock \bibinfo{title}{Tracking recurring contexts using ensemble classifiers: an application to email filtering}.
\newblock \bibinfo{journal}{Knowledge and Information Systems} \bibinfo{volume}{22}, \bibinfo{pages}{371--391}.
%Type = Inproceedings
\bibitem[{Katzir et~al.(2021)Katzir, Elidan and El-Yaniv}]{katzir2021net}
\bibinfo{author}{Katzir, O.}, \bibinfo{author}{Elidan, G.}, \bibinfo{author}{El-Yaniv, R.}, \bibinfo{year}{2021}.
\newblock \bibinfo{title}{Net-dnf: Effective deep modeling of tabular data}, in: \bibinfo{booktitle}{International Conference on Learning Representations}.
%Type = Article
\bibitem[{Kemker et~al.(2018)Kemker, McClure, Abitino, Hayes and Kanan}]{kemker2018measuring}
\bibinfo{author}{Kemker, R.}, \bibinfo{author}{McClure, M.}, \bibinfo{author}{Abitino, A.}, \bibinfo{author}{Hayes, T.L.}, \bibinfo{author}{Kanan, C.}, \bibinfo{year}{2018}.
\newblock \bibinfo{title}{Measuring catastrophic forgetting in neural networks}.
\newblock \bibinfo{journal}{Proceedings of the AAAI Conference on Artificial Intelligence} \bibinfo{volume}{32}.
%Type = Inproceedings
\bibitem[{Keskar et~al.(2017)Keskar, Mudigere, Nocedal, Smelyanskiy and Tang}]{keskar2017large}
\bibinfo{author}{Keskar, N.S.}, \bibinfo{author}{Mudigere, D.}, \bibinfo{author}{Nocedal, J.}, \bibinfo{author}{Smelyanskiy, M.}, \bibinfo{author}{Tang, P.T.P.}, \bibinfo{year}{2017}.
\newblock \bibinfo{title}{On large-batch training for deep learning: Generalization gap and sharp minima}, in: \bibinfo{booktitle}{International Conference on Learning Representations}.
%Type = Article
\bibitem[{Kim et~al.(2022)Kim, Xiao, Konishi, Ke and Liu}]{lee2022theoretical}
\bibinfo{author}{Kim, G.}, \bibinfo{author}{Xiao, C.}, \bibinfo{author}{Konishi, T.}, \bibinfo{author}{Ke, Z.}, \bibinfo{author}{Liu, B.}, \bibinfo{year}{2022}.
\newblock \bibinfo{title}{A theoretical study on solving continual learning}.
\newblock \bibinfo{journal}{Advances in neural information processing systems} \bibinfo{volume}{35}, \bibinfo{pages}{5065--5079}.
%Type = Article
\bibitem[{Kirkpatrick et~al.(2017)Kirkpatrick, Pascanu, Rabinowitz, Veness, Desjardins, Rusu, Milan, Quan, Ramalho, Grabska-Barwinska et~al.}]{kirkpatrick2017overcoming}
\bibinfo{author}{Kirkpatrick, J.}, \bibinfo{author}{Pascanu, R.}, \bibinfo{author}{Rabinowitz, N.}, \bibinfo{author}{Veness, J.}, \bibinfo{author}{Desjardins, G.}, \bibinfo{author}{Rusu, A.A.}, \bibinfo{author}{Milan, K.}, \bibinfo{author}{Quan, J.}, \bibinfo{author}{Ramalho, T.}, \bibinfo{author}{Grabska-Barwinska, A.}, et~al., \bibinfo{year}{2017}.
\newblock \bibinfo{title}{Overcoming catastrophic forgetting in neural networks}.
\newblock \bibinfo{journal}{Proceedings of the national academy of sciences} \bibinfo{volume}{114}, \bibinfo{pages}{3521--3526}.
%Type = Inproceedings
\bibitem[{Kirsch et~al.(2018)Kirsch, Kunze and Barber}]{kirsch2018modular}
\bibinfo{author}{Kirsch, L.}, \bibinfo{author}{Kunze, J.}, \bibinfo{author}{Barber, D.}, \bibinfo{year}{2018}.
\newblock \bibinfo{title}{Modular networks: Learning to decompose neural computation}, in: \bibinfo{booktitle}{Advances in Neural Information Processing Systems}, pp. \bibinfo{pages}{2408--2418}.
%Type = Article
\bibitem[{Knoblauch et~al.(2020)Knoblauch, Husain and Diethe}]{knoblauch2020optimal}
\bibinfo{author}{Knoblauch, J.}, \bibinfo{author}{Husain, H.}, \bibinfo{author}{Diethe, T.}, \bibinfo{year}{2020}.
\newblock \bibinfo{title}{Optimal continual learning has perfect memory and is np-hard}.
\newblock \bibinfo{journal}{arXiv preprint arXiv:2006.05188} .
%Type = Inproceedings
\bibitem[{Kornblith et~al.(2019)Kornblith, Norouzi, Lee and Hinton}]{kornblith2019similarity}
\bibinfo{author}{Kornblith, S.}, \bibinfo{author}{Norouzi, M.}, \bibinfo{author}{Lee, H.}, \bibinfo{author}{Hinton, G.}, \bibinfo{year}{2019}.
\newblock \bibinfo{title}{Similarity of neural network representations revisited}, in: \bibinfo{booktitle}{International Conference on Machine Learning}, \bibinfo{organization}{PMLR}. pp. \bibinfo{pages}{3519--3529}.
%Type = Article
\bibitem[{Kreps(2014)}]{kreps2014questioning}
\bibinfo{author}{Kreps, J.}, \bibinfo{year}{2014}.
\newblock \bibinfo{title}{Questioning the lambda architecture. the lambda architecture has its merits, but alternatives are worth exploring}.
\newblock \bibinfo{journal}{O'Reilly Media} \URLprefix \url{https://www.oreilly.com/ideas/questioning-the-lambda-architecture}.
%Type = Article
\bibitem[{Kumar et~al.(2023)Kumar, Marklund and Van~Roy}]{kumar2023maintaining}
\bibinfo{author}{Kumar, S.}, \bibinfo{author}{Marklund, H.}, \bibinfo{author}{Van~Roy, B.}, \bibinfo{year}{2023}.
\newblock \bibinfo{title}{Maintaining plasticity via regenerative regularization}.
\newblock \bibinfo{journal}{arXiv preprint arXiv:2308.11958} .
%Type = Article
\bibitem[{Lefa et~al.(2022)Lefa, Hatem and Salem}]{lefa2022enhancement}
\bibinfo{author}{Lefa, M.}, \bibinfo{author}{Hatem, A.}, \bibinfo{author}{Salem, R.}, \bibinfo{year}{2022}.
\newblock \bibinfo{title}{Enhancement of very fast decision tree for data stream mining}.
\newblock \bibinfo{journal}{Studies in Informatics and Control} \bibinfo{volume}{31}, \bibinfo{pages}{49--60}.
%Type = Article
\bibitem[{Lewandowski et~al.(2023)Lewandowski, Tanaka, Schuurmans and Machado}]{lewandowski2023curvature}
\bibinfo{author}{Lewandowski, A.}, \bibinfo{author}{Tanaka, H.}, \bibinfo{author}{Schuurmans, D.}, \bibinfo{author}{Machado, M.C.}, \bibinfo{year}{2023}.
\newblock \bibinfo{title}{Curvature explains loss of plasticity} .
%Type = Article
\bibitem[{Li et~al.(2016)Li, Kadav, Durdanovic, Samet and Graf}]{li2016pruning}
\bibinfo{author}{Li, H.}, \bibinfo{author}{Kadav, A.}, \bibinfo{author}{Durdanovic, I.}, \bibinfo{author}{Samet, H.}, \bibinfo{author}{Graf, H.P.}, \bibinfo{year}{2016}.
\newblock \bibinfo{title}{Pruning filters for efficient convnets}.
\newblock \bibinfo{journal}{arXiv preprint arXiv:1608.08710} .
%Type = Article
\bibitem[{Lin(2017)}]{lin2017lambda}
\bibinfo{author}{Lin, J.}, \bibinfo{year}{2017}.
\newblock \bibinfo{title}{The lambda and the kappa}.
\newblock \bibinfo{journal}{IEEE Internet Computing} \bibinfo{volume}{21}, \bibinfo{pages}{60--66}.
%Type = Article
\bibitem[{Lindstrom et~al.(2013)Lindstrom, Mac~Namee and Delany}]{lindstrom2013drift}
\bibinfo{author}{Lindstrom, P.}, \bibinfo{author}{Mac~Namee, B.}, \bibinfo{author}{Delany, S.J.}, \bibinfo{year}{2013}.
\newblock \bibinfo{title}{Drift detection using uncertainty distribution divergence}.
\newblock \bibinfo{journal}{Evolving Systems} \bibinfo{volume}{4}, \bibinfo{pages}{13--25}.
%Type = Inproceedings
\bibitem[{Liu et~al.(2018)Liu, Simonyan and Yang}]{liu2018darts}
\bibinfo{author}{Liu, H.}, \bibinfo{author}{Simonyan, K.}, \bibinfo{author}{Yang, Y.}, \bibinfo{year}{2018}.
\newblock \bibinfo{title}{Darts: Differentiable architecture search}, in: \bibinfo{booktitle}{International Conference on Learning Representations}.
%Type = Article
\bibitem[{Liu et~al.(2023)Liu, Biderman, Schoelkopf, Downey, Prashanth, Ramasesh and Schmidt}]{liu2023small}
\bibinfo{author}{Liu, S.}, \bibinfo{author}{Biderman, S.}, \bibinfo{author}{Schoelkopf, H.}, \bibinfo{author}{Downey, C.}, \bibinfo{author}{Prashanth, R.}, \bibinfo{author}{Ramasesh, V.V.}, \bibinfo{author}{Schmidt, L.}, \bibinfo{year}{2023}.
\newblock \bibinfo{title}{Small-scale proxies for large-scale transformer training instabilities}.
\newblock \bibinfo{journal}{arXiv preprint arXiv:2305.14718} .
%Type = Inproceedings
\bibitem[{Liu et~al.(2017)Liu, Li, Shen, Huang, Yan and Zhang}]{liu2017learning}
\bibinfo{author}{Liu, Z.}, \bibinfo{author}{Li, J.}, \bibinfo{author}{Shen, Z.}, \bibinfo{author}{Huang, G.}, \bibinfo{author}{Yan, S.}, \bibinfo{author}{Zhang, C.}, \bibinfo{year}{2017}.
\newblock \bibinfo{title}{Learning efficient convolutional networks through network slimming}, in: \bibinfo{booktitle}{Proceedings of the IEEE international conference on computer vision}, pp. \bibinfo{pages}{2736--2744}.
%Type = Inproceedings
\bibitem[{Liu et~al.(2019)Liu, Sun, Zhou, Huang and Darrell}]{liu2019rethinking}
\bibinfo{author}{Liu, Z.}, \bibinfo{author}{Sun, M.}, \bibinfo{author}{Zhou, T.}, \bibinfo{author}{Huang, G.}, \bibinfo{author}{Darrell, T.}, \bibinfo{year}{2019}.
\newblock \bibinfo{title}{Rethinking the value of network pruning}, in: \bibinfo{booktitle}{International Conference on Learning Representations}.
%Type = Inproceedings
\bibitem[{Losing et~al.(2018)Losing, Wersing and Hammer}]{losing2018enhancing}
\bibinfo{author}{Losing, V.}, \bibinfo{author}{Wersing, H.}, \bibinfo{author}{Hammer, B.}, \bibinfo{year}{2018}.
\newblock \bibinfo{title}{Enhancing very fast decision trees with local split-time predictions}, in: \bibinfo{booktitle}{2018 IEEE international conference on data mining (ICDM)}, \bibinfo{organization}{IEEE}. pp. \bibinfo{pages}{287--296}.
%Type = Article
\bibitem[{Louizos et~al.(2017)Louizos, Welling and Kingma}]{louizos2017learning}
\bibinfo{author}{Louizos, C.}, \bibinfo{author}{Welling, M.}, \bibinfo{author}{Kingma, D.P.}, \bibinfo{year}{2017}.
\newblock \bibinfo{title}{Learning sparse neural networks through $l_0$ regularization}.
\newblock \bibinfo{journal}{arXiv preprint arXiv:1712.01312} .
%Type = Article
\bibitem[{Louren\c{c}o et~al.(2025a)Louren\c{c}o, Gama, Xing and Marreiros}]{lourenco2025incontext}
\bibinfo{author}{Louren\c{c}o, A.}, \bibinfo{author}{Gama, J.a.}, \bibinfo{author}{Xing, E.P.}, \bibinfo{author}{Marreiros, G.}, \bibinfo{year}{2025}a.
\newblock \bibinfo{title}{In-context learning of evolving data streams with tabular foundational models}.
\newblock \bibinfo{journal}{arXiv preprint arXiv:2502.16840} .
%Type = Article
\bibitem[{Louren\c{c}o et~al.(2025b)Louren\c{c}o, Rodrigo, Gama and Marreiros}]{lourenco2025dfdt}
\bibinfo{author}{Louren\c{c}o, A.}, \bibinfo{author}{Rodrigo, J.a.}, \bibinfo{author}{Gama, J.a.}, \bibinfo{author}{Marreiros, G.}, \bibinfo{year}{2025}b.
\newblock \bibinfo{title}{Dfdt: Dynamic fast decision tree for iot data stream mining on edge devices}.
\newblock \bibinfo{journal}{arXiv preprint arXiv:2502.14011} .
%Type = Article
\bibitem[{Lyle et~al.(2024)Lyle, Schuurmans, Gal and Machado}]{lyle2024disentangling}
\bibinfo{author}{Lyle, C.}, \bibinfo{author}{Schuurmans, D.}, \bibinfo{author}{Gal, Y.}, \bibinfo{author}{Machado, M.C.}, \bibinfo{year}{2024}.
\newblock \bibinfo{title}{Disentangling the causes of plasticity loss in neural networks}.
\newblock \bibinfo{journal}{arXiv preprint arXiv:2402.18762} .
%Type = Inproceedings
\bibitem[{Lyle et~al.(2023)Lyle, Zheng, Nikishin, Pires, Pascanu and Dabney}]{lyle2023understanding}
\bibinfo{author}{Lyle, C.}, \bibinfo{author}{Zheng, Z.}, \bibinfo{author}{Nikishin, E.}, \bibinfo{author}{Pires, B.A.}, \bibinfo{author}{Pascanu, R.}, \bibinfo{author}{Dabney, W.}, \bibinfo{year}{2023}.
\newblock \bibinfo{title}{Understanding plasticity in neural networks}, in: \bibinfo{booktitle}{International Conference on Machine Learning}, \bibinfo{organization}{PMLR}. pp. \bibinfo{pages}{23190--23211}.
%Type = Inproceedings
\bibitem[{Manapragada et~al.(2018)Manapragada, Webb and Salehi}]{manapragada2018extremely}
\bibinfo{author}{Manapragada, C.}, \bibinfo{author}{Webb, G.I.}, \bibinfo{author}{Salehi, M.}, \bibinfo{year}{2018}.
\newblock \bibinfo{title}{Extremely fast decision tree}, in: \bibinfo{booktitle}{Proceedings of the 24th ACM SIGKDD International Conference on Knowledge Discovery \& Data Mining}, pp. \bibinfo{pages}{1953--1962}.
%Type = Article
\bibitem[{Masud et~al.(2013)Masud, Chen, Khan, Aggarwal, Gao, Han, Srivastava and Oza}]{masud2013classification}
\bibinfo{author}{Masud, M.M.}, \bibinfo{author}{Chen, Q.}, \bibinfo{author}{Khan, L.}, \bibinfo{author}{Aggarwal, C.C.}, \bibinfo{author}{Gao, J.}, \bibinfo{author}{Han, J.}, \bibinfo{author}{Srivastava, A.}, \bibinfo{author}{Oza, N.C.}, \bibinfo{year}{2013}.
\newblock \bibinfo{title}{Classification and novel class detection of data streams in a dynamic feature space}.
\newblock \bibinfo{journal}{IEEE Transactions on Knowledge and Data Engineering} \bibinfo{volume}{26}, \bibinfo{pages}{965--979}.
%Type = Inproceedings
\bibitem[{Masud et~al.(2011)Masud, Gao, Khan, Han and Thuraisingham}]{masud2011classification}
\bibinfo{author}{Masud, M.M.}, \bibinfo{author}{Gao, J.}, \bibinfo{author}{Khan, L.}, \bibinfo{author}{Han, J.}, \bibinfo{author}{Thuraisingham, B.}, \bibinfo{year}{2011}.
\newblock \bibinfo{title}{Classification and novel class detection in concept-drifting data streams under time constraints}, in: \bibinfo{booktitle}{IEEE Transactions on Knowledge and Data Engineering}, \bibinfo{organization}{IEEE}. pp. \bibinfo{pages}{859--874}.
%Type = Article
\bibitem[{Merrill et~al.(2023)Merrill, Goldberg, Schwartz and Smith}]{merrill2023effects}
\bibinfo{author}{Merrill, W.}, \bibinfo{author}{Goldberg, Y.}, \bibinfo{author}{Schwartz, R.}, \bibinfo{author}{Smith, N.A.}, \bibinfo{year}{2023}.
\newblock \bibinfo{title}{Effects of parameter norm growth during transformer training: Inductive bias from gradient descent}.
\newblock \bibinfo{journal}{Transactions of the Association for Computational Linguistics} \bibinfo{volume}{11}, \bibinfo{pages}{121--136}.
%Type = Article
\bibitem[{Mirkhan et~al.(2021)Mirkhan, Amir~Haeri and Meybodi}]{mirkhan2021analytical}
\bibinfo{author}{Mirkhan, M.}, \bibinfo{author}{Amir~Haeri, M.}, \bibinfo{author}{Meybodi, M.R.}, \bibinfo{year}{2021}.
\newblock \bibinfo{title}{Analytical split value calculation for numerical attributes in hoeffding trees with misclassification-based impurity}.
\newblock \bibinfo{journal}{Annals of Data Science} \bibinfo{volume}{8}, \bibinfo{pages}{645--665}.
%Type = Inproceedings
\bibitem[{Mirzadeh et~al.(2020)Mirzadeh, Farajtabar, Pascanu and Ghasemzadeh}]{mirzadeh2020understanding}
\bibinfo{author}{Mirzadeh, S.I.}, \bibinfo{author}{Farajtabar, M.}, \bibinfo{author}{Pascanu, R.}, \bibinfo{author}{Ghasemzadeh, H.}, \bibinfo{year}{2020}.
\newblock \bibinfo{title}{Understanding the role of training regimes in continual learning}, in: \bibinfo{booktitle}{Advances in Neural Information Processing Systems}.
%Type = Article
\bibitem[{Mirzadeh et~al.(2023)Mirzadeh, Farajtabar, Pascanu and Ghasemzadeh}]{mirzadeh2023directions}
\bibinfo{author}{Mirzadeh, S.I.}, \bibinfo{author}{Farajtabar, M.}, \bibinfo{author}{Pascanu, R.}, \bibinfo{author}{Ghasemzadeh, H.}, \bibinfo{year}{2023}.
\newblock \bibinfo{title}{Directions of curvature as an explanation for loss of plasticity}.
\newblock \bibinfo{journal}{arXiv preprint arXiv:2302.03044} .
%Type = Inproceedings
\bibitem[{Molchanov et~al.(2019)Molchanov, Mallya, Tyree, Frosio and Kautz}]{molchanov2019importance}
\bibinfo{author}{Molchanov, P.}, \bibinfo{author}{Mallya, A.}, \bibinfo{author}{Tyree, S.}, \bibinfo{author}{Frosio, I.}, \bibinfo{author}{Kautz, J.}, \bibinfo{year}{2019}.
\newblock \bibinfo{title}{Importance estimation for neural network pruning}, in: \bibinfo{booktitle}{Proceedings of the IEEE/CVF Conference on Computer Vision and Pattern Recognition}, pp. \bibinfo{pages}{11264--11272}.
%Type = Article
\bibitem[{Molchanov et~al.(2016)Molchanov, Tyree, Karras, Aila and Kautz}]{molchanov2016pruning}
\bibinfo{author}{Molchanov, P.}, \bibinfo{author}{Tyree, S.}, \bibinfo{author}{Karras, T.}, \bibinfo{author}{Aila, T.}, \bibinfo{author}{Kautz, J.}, \bibinfo{year}{2016}.
\newblock \bibinfo{title}{Pruning convolutional neural networks for resource efficient inference}.
\newblock \bibinfo{journal}{arXiv preprint arXiv:1611.06440} .
%Type = Article
\bibitem[{Mouss et~al.(2004)Mouss, Mouss, Mouss and Sefouhi}]{mouss2004test}
\bibinfo{author}{Mouss, H.}, \bibinfo{author}{Mouss, D.}, \bibinfo{author}{Mouss, N.}, \bibinfo{author}{Sefouhi, L.}, \bibinfo{year}{2004}.
\newblock \bibinfo{title}{Test of page-hinckley, an approach for fault detection in an agro-alimentary production system}.
\newblock \bibinfo{journal}{2004 5th Asian Control Conference (IEEE Cat. No. 04EX904)} \bibinfo{volume}{2}, \bibinfo{pages}{815--818}.
%Type = Inproceedings
\bibitem[{Nagel et~al.(2019)Nagel, Baalen, Blankevoort and Welling}]{nagel2019data}
\bibinfo{author}{Nagel, M.}, \bibinfo{author}{Baalen, M.v.}, \bibinfo{author}{Blankevoort, T.}, \bibinfo{author}{Welling, M.}, \bibinfo{year}{2019}.
\newblock \bibinfo{title}{Data-free quantization through weight equalization and bias correction}, in: \bibinfo{booktitle}{Proceedings of the IEEE/CVF International Conference on Computer Vision}, pp. \bibinfo{pages}{1325--1334}.
%Type = Inproceedings
\bibitem[{Nguyen et~al.(2018)Nguyen, Li, Bui and Turner}]{nguyen2018variational}
\bibinfo{author}{Nguyen, C.V.}, \bibinfo{author}{Li, Y.}, \bibinfo{author}{Bui, T.D.}, \bibinfo{author}{Turner, R.E.}, \bibinfo{year}{2018}.
\newblock \bibinfo{title}{Variational continual learning}, in: \bibinfo{booktitle}{International Conference on Learning Representations}.
%Type = Inproceedings
\bibitem[{Nikishin et~al.(2024)Nikishin, Oh, Ostrovski, Lyle, Pascanu, Dabney and Barreto}]{nikishin2024deep}
\bibinfo{author}{Nikishin, E.}, \bibinfo{author}{Oh, J.}, \bibinfo{author}{Ostrovski, G.}, \bibinfo{author}{Lyle, C.}, \bibinfo{author}{Pascanu, R.}, \bibinfo{author}{Dabney, W.}, \bibinfo{author}{Barreto, A.}, \bibinfo{year}{2024}.
\newblock \bibinfo{title}{Deep reinforcement learning with plasticity injection}, in: \bibinfo{booktitle}{Advances in Neural Information Processing Systems}.
%Type = Article
\bibitem[{N{\'u}{\~n}ez et~al.(2007)N{\'u}{\~n}ez, Fidalgo and Morales}]{nunez2007learning}
\bibinfo{author}{N{\'u}{\~n}ez, M.}, \bibinfo{author}{Fidalgo, R.}, \bibinfo{author}{Morales, R.}, \bibinfo{year}{2007}.
\newblock \bibinfo{title}{Learning in environments with unknown dynamics: Towards more robust concept learners.}
\newblock \bibinfo{journal}{Journal of Machine Learning Research} \bibinfo{volume}{8}.
%Type = Inproceedings
\bibitem[{Ostapenko et~al.(2022)Ostapenko, Suris, Szab{\'o} and Mikolov}]{ostapenko2022attention}
\bibinfo{author}{Ostapenko, O.}, \bibinfo{author}{Suris, D.}, \bibinfo{author}{Szab{\'o}, A.}, \bibinfo{author}{Mikolov, T.}, \bibinfo{year}{2022}.
\newblock \bibinfo{title}{Attention for compositional modularity}, in: \bibinfo{booktitle}{NeurIPS'22 Workshop on All Things Attention: Bridging Different Perspectives on Attention}.
%Type = Article
\bibitem[{von Oswald et~al.(2019)von Oswald, Henning, Sacramento and Grewe}]{von2019continual}
\bibinfo{author}{von Oswald, J.}, \bibinfo{author}{Henning, C.}, \bibinfo{author}{Sacramento, J.}, \bibinfo{author}{Grewe, B.F.}, \bibinfo{year}{2019}.
\newblock \bibinfo{title}{Continual learning with hypernetworks}.
\newblock \bibinfo{journal}{arXiv preprint arXiv:1906.00695} .
%Type = Inproceedings
\bibitem[{Oza and Russell(2001)}]{oza2001experimental}
\bibinfo{author}{Oza, N.C.}, \bibinfo{author}{Russell, S.}, \bibinfo{year}{2001}.
\newblock \bibinfo{title}{Experimental comparisons of online and batch versions of bagging and boosting}, in: \bibinfo{booktitle}{Proceedings of the seventh ACM SIGKDD international conference on Knowledge discovery and data mining}, pp. \bibinfo{pages}{359--364}.
%Type = Inproceedings
\bibitem[{Park et~al.(2020)Park, Yoo, Cho, Kim and Yoon}]{park2020lookahead}
\bibinfo{author}{Park, S.}, \bibinfo{author}{Yoo, J.}, \bibinfo{author}{Cho, D.}, \bibinfo{author}{Kim, J.}, \bibinfo{author}{Yoon, S.}, \bibinfo{year}{2020}.
\newblock \bibinfo{title}{Lookahead: A far-sighted alternative of magnitude-based pruning}, in: \bibinfo{booktitle}{International Conference on Learning Representations}.
%Type = Article
\bibitem[{Pietruczuk et~al.(2017)Pietruczuk, Rutkowski, Jaworski and Duda}]{pietruczuk2017automatic}
\bibinfo{author}{Pietruczuk, L.}, \bibinfo{author}{Rutkowski, L.}, \bibinfo{author}{Jaworski, M.}, \bibinfo{author}{Duda, P.}, \bibinfo{year}{2017}.
\newblock \bibinfo{title}{A method for automatic adjustment of ensemble size in stream data mining}.
\newblock \bibinfo{journal}{Information Sciences} \bibinfo{volume}{381}, \bibinfo{pages}{46--54}.
%Type = Article
\bibitem[{Popov et~al.(2019)Popov, Morozov and Babenko}]{popov2019neural}
\bibinfo{author}{Popov, S.}, \bibinfo{author}{Morozov, S.}, \bibinfo{author}{Babenko, A.}, \bibinfo{year}{2019}.
\newblock \bibinfo{title}{Neural oblivious decision ensembles for deep learning on tabular data}.
\newblock \bibinfo{journal}{arXiv preprint arXiv:1909.06312} .
%Type = Inproceedings
\bibitem[{Prabhu et~al.(2020)Prabhu, Torr and Dokania}]{prabhu2020gdumb}
\bibinfo{author}{Prabhu, A.}, \bibinfo{author}{Torr, P.H.}, \bibinfo{author}{Dokania, P.K.}, \bibinfo{year}{2020}.
\newblock \bibinfo{title}{Gdumb: A simple approach that questions our progress in continual learning}, in: \bibinfo{booktitle}{European Conference on Computer Vision}, \bibinfo{organization}{Springer}. pp. \bibinfo{pages}{524--540}.
%Type = Inproceedings
\bibitem[{Raghu et~al.(2017)Raghu, Gilmer, Yosinski and Sohl-Dickstein}]{raghu2017svcca}
\bibinfo{author}{Raghu, M.}, \bibinfo{author}{Gilmer, J.}, \bibinfo{author}{Yosinski, J.}, \bibinfo{author}{Sohl-Dickstein, J.}, \bibinfo{year}{2017}.
\newblock \bibinfo{title}{Svcca: Singular vector canonical correlation analysis for deep learning dynamics and interpretability}, in: \bibinfo{booktitle}{Advances in Neural Information Processing Systems}, pp. \bibinfo{pages}{6076--6085}.
%Type = Article
\bibitem[{Ramasesh et~al.(2020a)Ramasesh, Dyer and Raghu}]{ramasesh2021anatomy}
\bibinfo{author}{Ramasesh, V.V.}, \bibinfo{author}{Dyer, E.}, \bibinfo{author}{Raghu, M.}, \bibinfo{year}{2020}a.
\newblock \bibinfo{title}{Anatomy of catastrophic forgetting: Hidden representations and task semantics}.
\newblock \bibinfo{journal}{arXiv preprint arXiv:2007.07400} .
%Type = Article
\bibitem[{Ramasesh et~al.(2020b)Ramasesh, Dyer and Raghu}]{ramasesh2020anatomy}
\bibinfo{author}{Ramasesh, V.V.}, \bibinfo{author}{Dyer, E.}, \bibinfo{author}{Raghu, M.}, \bibinfo{year}{2020}b.
\newblock \bibinfo{title}{Anatomy of catastrophic forgetting: Hidden representations and task semantics}.
\newblock \bibinfo{journal}{arXiv preprint arXiv:2007.07400} .
%Type = Article
\bibitem[{Razavi-Far et~al.(2018)Razavi-Far, Hallaji, Saif and Ditzler}]{costa2019novelty}
\bibinfo{author}{Razavi-Far, R.}, \bibinfo{author}{Hallaji, E.}, \bibinfo{author}{Saif, M.}, \bibinfo{author}{Ditzler, G.}, \bibinfo{year}{2018}.
\newblock \bibinfo{title}{A novelty detector and extreme verification latency model for nonstationary environments}.
\newblock \bibinfo{journal}{IEEE Transactions on Industrial Electronics} \bibinfo{volume}{66}, \bibinfo{pages}{561--570}.
%Type = Article
\bibitem[{Razin and Cohen(2020)}]{razin2020implicit}
\bibinfo{author}{Razin, N.}, \bibinfo{author}{Cohen, N.}, \bibinfo{year}{2020}.
\newblock \bibinfo{title}{Implicit regularization in deep learning may not be explainable by norms}.
\newblock \bibinfo{journal}{Advances in Neural Information Processing Systems} \bibinfo{volume}{33}, \bibinfo{pages}{2803--2814}.
%Type = Inproceedings
\bibitem[{dos Reis et~al.(2016)dos Reis, Flach, Matwin and Batista}]{dos2016fast}
\bibinfo{author}{dos Reis, D.M.}, \bibinfo{author}{Flach, P.}, \bibinfo{author}{Matwin, S.}, \bibinfo{author}{Batista, G.}, \bibinfo{year}{2016}.
\newblock \bibinfo{title}{Fast unsupervised online drift detection using incremental kolmogorov-smirnov test}, in: \bibinfo{booktitle}{Proceedings of the 22nd ACM SIGKDD International Conference on Knowledge Discovery and Data Mining}, pp. \bibinfo{pages}{1545--1554}.
%Type = Article
\bibitem[{Rosenbaum et~al.(2019)Rosenbaum, Klinger and Riemer}]{rosenbaum2019routing}
\bibinfo{author}{Rosenbaum, C.}, \bibinfo{author}{Klinger, T.}, \bibinfo{author}{Riemer, M.}, \bibinfo{year}{2019}.
\newblock \bibinfo{title}{Routing networks and the challenges of modular and compositional computation}.
\newblock \bibinfo{journal}{arXiv preprint arXiv:1904.12774} .
%Type = Inproceedings
\bibitem[{Roy and Vetterli(2007)}]{roy2007effective}
\bibinfo{author}{Roy, O.}, \bibinfo{author}{Vetterli, M.}, \bibinfo{year}{2007}.
\newblock \bibinfo{title}{The effective rank: A measure of effective dimensionality}, in: \bibinfo{booktitle}{2007 15th European Signal Processing Conference}, \bibinfo{organization}{IEEE}. pp. \bibinfo{pages}{606--610}.
%Type = Article
\bibitem[{Rusu et~al.(2016)Rusu, Rabinowitz, Desjardins, Soyer, Kirkpatrick, Kavukcuoglu, Pascanu and Hadsell}]{rusu2016progressive}
\bibinfo{author}{Rusu, A.A.}, \bibinfo{author}{Rabinowitz, N.C.}, \bibinfo{author}{Desjardins, G.}, \bibinfo{author}{Soyer, H.}, \bibinfo{author}{Kirkpatrick, J.}, \bibinfo{author}{Kavukcuoglu, K.}, \bibinfo{author}{Pascanu, R.}, \bibinfo{author}{Hadsell, R.}, \bibinfo{year}{2016}.
\newblock \bibinfo{title}{Progressive neural networks}.
\newblock \bibinfo{journal}{arXiv preprint arXiv:1606.04671} .
%Type = Article
\bibitem[{Rutkowski et~al.(2014)Rutkowski, Jaworski, Pietruczuk and Duda}]{rutkowski2014new}
\bibinfo{author}{Rutkowski, L.}, \bibinfo{author}{Jaworski, M.}, \bibinfo{author}{Pietruczuk, L.}, \bibinfo{author}{Duda, P.}, \bibinfo{year}{2014}.
\newblock \bibinfo{title}{A new method for data stream mining based on the misclassification error}.
\newblock \bibinfo{journal}{IEEE transactions on neural networks and learning systems} \bibinfo{volume}{26}, \bibinfo{pages}{1048--1059}.
%Type = Article
\bibitem[{Rutkowski et~al.(2012)Rutkowski, Pietruczuk, Duda and Jaworski}]{rutkowski2012decision}
\bibinfo{author}{Rutkowski, L.}, \bibinfo{author}{Pietruczuk, L.}, \bibinfo{author}{Duda, P.}, \bibinfo{author}{Jaworski, M.}, \bibinfo{year}{2012}.
\newblock \bibinfo{title}{Decision trees for mining data streams based on the mcdiarmid's bound}.
\newblock \bibinfo{journal}{IEEE Transactions on Knowledge and Data Engineering} \bibinfo{volume}{25}, \bibinfo{pages}{1272--1279}.
%Type = Article
\bibitem[{Salehi et~al.(2021)Salehi, Sadjadi, Baghshah, Rabiee and Rohban}]{salehi2021unified}
\bibinfo{author}{Salehi, M.}, \bibinfo{author}{Sadjadi, N.}, \bibinfo{author}{Baghshah, S.}, \bibinfo{author}{Rabiee, H.R.}, \bibinfo{author}{Rohban, M.H.}, \bibinfo{year}{2021}.
\newblock \bibinfo{title}{A unified survey on anomaly, novelty, open-set, and out-of-distribution detection: Solutions and future challenges}.
\newblock \bibinfo{journal}{arXiv preprint arXiv:2110.14051} .
%Type = Article
\bibitem[{Schizas et~al.(2022)Schizas, Karras, Karras and Sioutas}]{schizas2022tinyml}
\bibinfo{author}{Schizas, N.}, \bibinfo{author}{Karras, A.}, \bibinfo{author}{Karras, C.}, \bibinfo{author}{Sioutas, S.}, \bibinfo{year}{2022}.
\newblock \bibinfo{title}{Tinyml for ultra-low power ai and large scale iot deployments: A systematic review}.
\newblock \bibinfo{journal}{Future Internet} \bibinfo{volume}{14}, \bibinfo{pages}{363}.
%Type = Article
\bibitem[{Schwarz et~al.(2018)Schwarz, Luketina, Czarnecki, Grabska-Barwinska, Teh, Pascanu and Hadsell}]{schwarz2018progress}
\bibinfo{author}{Schwarz, J.}, \bibinfo{author}{Luketina, J.}, \bibinfo{author}{Czarnecki, W.M.}, \bibinfo{author}{Grabska-Barwinska, A.}, \bibinfo{author}{Teh, Y.W.}, \bibinfo{author}{Pascanu, R.}, \bibinfo{author}{Hadsell, R.}, \bibinfo{year}{2018}.
\newblock \bibinfo{title}{Progress \& compress: A scalable framework for continual learning}.
\newblock \bibinfo{journal}{International Conference on Machine Learning} , \bibinfo{pages}{4528--4537}.
%Type = Article
\bibitem[{Serra et~al.(2018)Serra, Suris, Miron and Karatzoglou}]{serra2018overcoming}
\bibinfo{author}{Serra, J.}, \bibinfo{author}{Suris, D.}, \bibinfo{author}{Miron, M.}, \bibinfo{author}{Karatzoglou, A.}, \bibinfo{year}{2018}.
\newblock \bibinfo{title}{Overcoming catastrophic forgetting with hard attention to the task}.
\newblock \bibinfo{journal}{International Conference on Machine Learning} , \bibinfo{pages}{4548--4557}.
%Type = Article
\bibitem[{Sethi and Kantardzic(2015)}]{sethi2016md3}
\bibinfo{author}{Sethi, T.S.}, \bibinfo{author}{Kantardzic, M.}, \bibinfo{year}{2015}.
\newblock \bibinfo{title}{Don’t pay for validation: Detecting drifts from unlabeled data using margin density}.
\newblock \bibinfo{journal}{Procedia Computer Science} \bibinfo{volume}{53}, \bibinfo{pages}{103--112}.
%Type = Article
\bibitem[{Shaker and H{\"u}llermeier(2015)}]{shaker2015recovery}
\bibinfo{author}{Shaker, A.}, \bibinfo{author}{H{\"u}llermeier, E.}, \bibinfo{year}{2015}.
\newblock \bibinfo{title}{Recovery analysis for adaptive learning from non-stationary data streams: Experimental design and case study}.
\newblock \bibinfo{journal}{Neurocomputing} \bibinfo{volume}{150}, \bibinfo{pages}{250--264}.
%Type = Inproceedings
\bibitem[{Shazeer et~al.(2017)Shazeer, Mirhoseini, Maziarz, Davis, Le, Hinton and Dean}]{shazeer2017outrageously}
\bibinfo{author}{Shazeer, N.}, \bibinfo{author}{Mirhoseini, A.}, \bibinfo{author}{Maziarz, K.}, \bibinfo{author}{Davis, A.}, \bibinfo{author}{Le, Q.}, \bibinfo{author}{Hinton, G.}, \bibinfo{author}{Dean, J.}, \bibinfo{year}{2017}.
\newblock \bibinfo{title}{Outrageously large neural networks: The sparsely-gated mixture-of-experts layer}, in: \bibinfo{booktitle}{International Conference on Learning Representations}.
%Type = Inproceedings
\bibitem[{Shin et~al.(2017)Shin, Lee, Kim and Kim}]{shin2017continual}
\bibinfo{author}{Shin, H.}, \bibinfo{author}{Lee, J.K.}, \bibinfo{author}{Kim, J.}, \bibinfo{author}{Kim, J.}, \bibinfo{year}{2017}.
\newblock \bibinfo{title}{Continual learning with deep generative replay}, in: \bibinfo{booktitle}{Advances in neural information processing systems}, pp. \bibinfo{pages}{2990--2999}.
%Type = Article
\bibitem[{Shwartz-Ziv and Armon(2022)}]{shwartz2022tabular}
\bibinfo{author}{Shwartz-Ziv, R.}, \bibinfo{author}{Armon, A.}, \bibinfo{year}{2022}.
\newblock \bibinfo{title}{Tabular data: Deep learning is not all you need}.
\newblock \bibinfo{journal}{Information Fusion} \bibinfo{volume}{81}, \bibinfo{pages}{84--90}.
%Type = Inproceedings
\bibitem[{Sokar et~al.(2023)Sokar, Agarwal, Castro and Evci}]{sokar2023dormant}
\bibinfo{author}{Sokar, G.}, \bibinfo{author}{Agarwal, R.}, \bibinfo{author}{Castro, P.S.}, \bibinfo{author}{Evci, U.}, \bibinfo{year}{2023}.
\newblock \bibinfo{title}{The dormant neuron phenomenon in deep reinforcement learning}, in: \bibinfo{booktitle}{International Conference on Machine Learning}, \bibinfo{organization}{PMLR}. pp. \bibinfo{pages}{32145--32168}.
%Type = Article
\bibitem[{Somepalli et~al.(2021)Somepalli, Goldblum, Schwarzschild, Bruss and Goldstein}]{somepalli2021saint}
\bibinfo{author}{Somepalli, G.}, \bibinfo{author}{Goldblum, M.}, \bibinfo{author}{Schwarzschild, A.}, \bibinfo{author}{Bruss, C.B.}, \bibinfo{author}{Goldstein, T.}, \bibinfo{year}{2021}.
\newblock \bibinfo{title}{Saint: Improved neural networks for tabular data via row attention and contrastive pre-training}.
\newblock \bibinfo{journal}{arXiv preprint arXiv:2106.01342} .
%Type = Article
\bibitem[{Spinosa et~al.(2009)Spinosa, de~Carvalho and Gama}]{spinosa2009olindda}
\bibinfo{author}{Spinosa, E.J.}, \bibinfo{author}{de~Carvalho, A.C.}, \bibinfo{author}{Gama, J.}, \bibinfo{year}{2009}.
\newblock \bibinfo{title}{Olindda: A cluster-based approach for detecting novelty and concept drift in data streams}.
\newblock \bibinfo{journal}{Expert Systems with Applications} \bibinfo{volume}{36}, \bibinfo{pages}{4370--4382}.
%Type = Article
\bibitem[{Sun et~al.(2020)Sun, Ren, Liang, Zhang, Jiang and Yin}]{sun2020survey}
\bibinfo{author}{Sun, H.}, \bibinfo{author}{Ren, X.}, \bibinfo{author}{Liang, S.}, \bibinfo{author}{Zhang, Y.}, \bibinfo{author}{Jiang, C.}, \bibinfo{author}{Yin, W.}, \bibinfo{year}{2020}.
\newblock \bibinfo{title}{A survey on dynamic neural networks for natural language processing}.
\newblock \bibinfo{journal}{arXiv preprint arXiv:2006.00693} .
%Type = Inproceedings
\bibitem[{Sun et~al.(2021)Sun, Jia, Hu, Huang, Zhang, Wan and Zhao}]{sun2021speeding}
\bibinfo{author}{Sun, J.}, \bibinfo{author}{Jia, H.}, \bibinfo{author}{Hu, B.}, \bibinfo{author}{Huang, X.}, \bibinfo{author}{Zhang, H.}, \bibinfo{author}{Wan, H.}, \bibinfo{author}{Zhao, X.}, \bibinfo{year}{2021}.
\newblock \bibinfo{title}{Speeding up very fast decision tree with low computational cost}, in: \bibinfo{booktitle}{Proceedings of the Twenty-Ninth International Conference on International Joint Conferences on Artificial Intelligence}, pp. \bibinfo{pages}{1272--7278}.
%Type = Inproceedings
\bibitem[{Tan et~al.(2020)Tan, Bergmeir, Petitjean and Webb}]{tan2020converting}
\bibinfo{author}{Tan, C.W.}, \bibinfo{author}{Bergmeir, C.}, \bibinfo{author}{Petitjean, F.}, \bibinfo{author}{Webb, G.I.}, \bibinfo{year}{2020}.
\newblock \bibinfo{title}{Converting tabular data into images for deep learning with convolutional neural networks}, in: \bibinfo{booktitle}{2020 International Joint Conference on Neural Networks (IJCNN)}, \bibinfo{organization}{IEEE}. pp. \bibinfo{pages}{1--8}.
%Type = Article
\bibitem[{Van~Dongen and Van~den Poel(2020)}]{van2020evaluation}
\bibinfo{author}{Van~Dongen, G.}, \bibinfo{author}{Van~den Poel, D.}, \bibinfo{year}{2020}.
\newblock \bibinfo{title}{Evaluation of stream processing frameworks}.
\newblock \bibinfo{journal}{IEEE Transactions on Parallel and Distributed Systems} \bibinfo{volume}{31}, \bibinfo{pages}{1845--1858}.
%Type = Article
\bibitem[{Verwimp et~al.(2023)Verwimp, Ben-David, Bethge, Cossu, Gepperth, Hayes, H{\"u}llermeier, Kanan, Kudithipudi, Lampert et~al.}]{verwimp2023continual}
\bibinfo{author}{Verwimp, E.}, \bibinfo{author}{Ben-David, S.}, \bibinfo{author}{Bethge, M.}, \bibinfo{author}{Cossu, A.}, \bibinfo{author}{Gepperth, A.}, \bibinfo{author}{Hayes, T.L.}, \bibinfo{author}{H{\"u}llermeier, E.}, \bibinfo{author}{Kanan, C.}, \bibinfo{author}{Kudithipudi, D.}, \bibinfo{author}{Lampert, C.H.}, et~al., \bibinfo{year}{2023}.
\newblock \bibinfo{title}{Continual learning: Applications and the road forward}.
\newblock \bibinfo{journal}{arXiv preprint arXiv:2311.11908} .
%Type = Article
\bibitem[{Wang et~al.(2018)Wang, Zhu, Torralba and Efros}]{wang2021dataset}
\bibinfo{author}{Wang, T.}, \bibinfo{author}{Zhu, J.Y.}, \bibinfo{author}{Torralba, A.}, \bibinfo{author}{Efros, A.A.}, \bibinfo{year}{2018}.
\newblock \bibinfo{title}{Dataset distillation}.
\newblock \bibinfo{journal}{arXiv preprint arXiv:1811.10959} .
%Type = Article
\bibitem[{Wang et~al.(2020)Wang, Yao, Kwok and Ni}]{wang2020generalizing}
\bibinfo{author}{Wang, Y.}, \bibinfo{author}{Yao, Q.}, \bibinfo{author}{Kwok, J.T.}, \bibinfo{author}{Ni, L.M.}, \bibinfo{year}{2020}.
\newblock \bibinfo{title}{Generalizing from a few examples: A survey on few-shot learning}.
\newblock \bibinfo{journal}{ACM Computing Surveys (CSUR)} \bibinfo{volume}{53}, \bibinfo{pages}{1--34}.
%Type = Article
\bibitem[{Wang and Sun(2022)}]{wang2022transtab}
\bibinfo{author}{Wang, Z.}, \bibinfo{author}{Sun, J.}, \bibinfo{year}{2022}.
\newblock \bibinfo{title}{Transtab: Learning transferable tabular transformers across tables}.
\newblock \bibinfo{journal}{Advances in Neural Information Processing Systems} \bibinfo{volume}{35}, \bibinfo{pages}{2902--2915}.
%Type = Book
\bibitem[{Warden and Situnayake(2019)}]{warden2019tinyml}
\bibinfo{author}{Warden, P.}, \bibinfo{author}{Situnayake, D.}, \bibinfo{year}{2019}.
\newblock \bibinfo{title}{Tinyml: Machine learning with tensorflow lite on arduino and ultra-low-power microcontrollers}.
\newblock \bibinfo{publisher}{O'Reilly Media}.
%Type = Book
\bibitem[{Warren and Marz(2015)}]{warren2015big}
\bibinfo{author}{Warren, J.}, \bibinfo{author}{Marz, N.}, \bibinfo{year}{2015}.
\newblock \bibinfo{title}{Big Data: Principles and best practices of scalable realtime data systems}.
\newblock \bibinfo{publisher}{Simon and Schuster}.
%Type = Inproceedings
\bibitem[{Wen et~al.(2016)Wen, Wu, Wang, Chen and Li}]{wen2016learning}
\bibinfo{author}{Wen, W.}, \bibinfo{author}{Wu, C.}, \bibinfo{author}{Wang, Y.}, \bibinfo{author}{Chen, Y.}, \bibinfo{author}{Li, H.}, \bibinfo{year}{2016}.
\newblock \bibinfo{title}{Learning structured sparsity in deep neural networks}, in: \bibinfo{booktitle}{Advances in neural information processing systems}.
%Type = Inproceedings
\bibitem[{Wong(2019)}]{wong2019netscore}
\bibinfo{author}{Wong, A.}, \bibinfo{year}{2019}.
\newblock \bibinfo{title}{Netscore: towards universal metrics for large-scale performance analysis of deep neural networks for practical on-device edge usage}, in: \bibinfo{booktitle}{International Conference on Image Analysis and Recognition}, \bibinfo{organization}{Springer}. pp. \bibinfo{pages}{15--26}.
%Type = Inproceedings
\bibitem[{Wortsman et~al.(2020)Wortsman, Ramanujan, Liu, Kembhavi, Rastegari, Yosinski and Farhadi}]{wortsman2020supermasks}
\bibinfo{author}{Wortsman, M.}, \bibinfo{author}{Ramanujan, V.}, \bibinfo{author}{Liu, R.}, \bibinfo{author}{Kembhavi, A.}, \bibinfo{author}{Rastegari, M.}, \bibinfo{author}{Yosinski, J.}, \bibinfo{author}{Farhadi, A.}, \bibinfo{year}{2020}.
\newblock \bibinfo{title}{Supermasks in superposition}, in: \bibinfo{booktitle}{Advances in Neural Information Processing Systems}, pp. \bibinfo{pages}{15173--15184}.
%Type = Article
\bibitem[{Wu et~al.(2021)Wu, Wu, Qi, Huang and Xie}]{wu2021fastformer}
\bibinfo{author}{Wu, C.}, \bibinfo{author}{Wu, F.}, \bibinfo{author}{Qi, T.}, \bibinfo{author}{Huang, Y.}, \bibinfo{author}{Xie, X.}, \bibinfo{year}{2021}.
\newblock \bibinfo{title}{Fastformer: Additive attention can be all you need}.
\newblock \bibinfo{journal}{arXiv preprint arXiv:2108.09084} .
%Type = Inproceedings
\bibitem[{Wu et~al.(2024)Wu, Chen, Zhao, Sergazinov, Li, Liu, Zhao, Xie, Guo, Ji et~al.}]{wu2024switchtab}
\bibinfo{author}{Wu, J.}, \bibinfo{author}{Chen, S.}, \bibinfo{author}{Zhao, Q.}, \bibinfo{author}{Sergazinov, R.}, \bibinfo{author}{Li, C.}, \bibinfo{author}{Liu, S.}, \bibinfo{author}{Zhao, C.}, \bibinfo{author}{Xie, T.}, \bibinfo{author}{Guo, H.}, \bibinfo{author}{Ji, C.}, et~al., \bibinfo{year}{2024}.
\newblock \bibinfo{title}{Switchtab: Switched autoencoders are effective tabular learners}, in: \bibinfo{booktitle}{Proceedings of the AAAI Conference on Artificial Intelligence}, pp. \bibinfo{pages}{15924--15933}.
%Type = Article
\bibitem[{Xian et~al.(2018)Xian, Lampert, Schiele and Akata}]{xian2018zero}
\bibinfo{author}{Xian, Y.}, \bibinfo{author}{Lampert, C.H.}, \bibinfo{author}{Schiele, B.}, \bibinfo{author}{Akata, Z.}, \bibinfo{year}{2018}.
\newblock \bibinfo{title}{Zero-shot learning—a comprehensive evaluation of the good, the bad and the ugly}.
\newblock \bibinfo{journal}{IEEE transactions on pattern analysis and machine intelligence} \bibinfo{volume}{41}, \bibinfo{pages}{2251--2265}.
%Type = Article
\bibitem[{Xu and Wang(2016)}]{xu2016mining}
\bibinfo{author}{Xu, S.}, \bibinfo{author}{Wang, J.}, \bibinfo{year}{2016}.
\newblock \bibinfo{title}{Mining recurring concept drifts with limited labeled streaming data}.
\newblock \bibinfo{journal}{ACM Transactions on Intelligent Systems and Technology (TIST)} \bibinfo{volume}{7}, \bibinfo{pages}{1--32}.
%Type = Inproceedings
\bibitem[{Yamada et~al.(2020)Yamada, Lindenbaum, Negahban and Kluger}]{yamada2020feature}
\bibinfo{author}{Yamada, Y.}, \bibinfo{author}{Lindenbaum, O.}, \bibinfo{author}{Negahban, S.}, \bibinfo{author}{Kluger, Y.}, \bibinfo{year}{2020}.
\newblock \bibinfo{title}{Feature selection using stochastic gates}, in: \bibinfo{booktitle}{International conference on machine learning}, \bibinfo{organization}{PMLR}. pp. \bibinfo{pages}{10648--10659}.
%Type = Inproceedings
\bibitem[{Yang and Fong(2011a)}]{yang2011moderated}
\bibinfo{author}{Yang, H.}, \bibinfo{author}{Fong, S.}, \bibinfo{year}{2011}a.
\newblock \bibinfo{title}{Moderated vfdt in stream mining using adaptive tie threshold and incremental pruning}, in: \bibinfo{booktitle}{Data Warehousing and Knowledge Discovery: 13th International Conference, DaWaK 2011, Toulouse, France, August 29-September 2, 2011. Proceedings 13}, \bibinfo{organization}{Springer}. pp. \bibinfo{pages}{471--483}.
%Type = Inproceedings
\bibitem[{Yang and Fong(2011b)}]{yang2011optimized}
\bibinfo{author}{Yang, H.}, \bibinfo{author}{Fong, S.}, \bibinfo{year}{2011}b.
\newblock \bibinfo{title}{Optimized very fast decision tree with balanced classification accuracy and compact tree size}, in: \bibinfo{booktitle}{The 3rd international conference on data mining and intelligent information technology applications}, \bibinfo{organization}{IEEE}. pp. \bibinfo{pages}{57--64}.
%Type = Article
\bibitem[{Ye et~al.(2024)Ye, Fan, Song, Zheng, Zhao, Guo and Chang}]{ye2024ptarl}
\bibinfo{author}{Ye, H.}, \bibinfo{author}{Fan, W.}, \bibinfo{author}{Song, X.}, \bibinfo{author}{Zheng, S.}, \bibinfo{author}{Zhao, H.}, \bibinfo{author}{Guo, D.}, \bibinfo{author}{Chang, Y.}, \bibinfo{year}{2024}.
\newblock \bibinfo{title}{Ptarl: Prototype-based tabular representation learning via space calibration}.
\newblock \bibinfo{journal}{arXiv preprint arXiv:2407.05364} .
%Type = Article
\bibitem[{Yoon et~al.(2018)Yoon, Yang, Lee and Hwang}]{yoon2018dynamically}
\bibinfo{author}{Yoon, J.}, \bibinfo{author}{Yang, E.}, \bibinfo{author}{Lee, J.}, \bibinfo{author}{Hwang, S.J.}, \bibinfo{year}{2018}.
\newblock \bibinfo{title}{Dynamically expandable networks}.
\newblock \bibinfo{journal}{arXiv preprint arXiv:1708.01547} .
%Type = Article
\bibitem[{Yoon et~al.(2020)Yoon, Zhang, Jordon and Van~der Schaar}]{yoon2020vime}
\bibinfo{author}{Yoon, J.}, \bibinfo{author}{Zhang, Y.}, \bibinfo{author}{Jordon, J.}, \bibinfo{author}{Van~der Schaar, M.}, \bibinfo{year}{2020}.
\newblock \bibinfo{title}{Vime: Extending the success of self-and semi-supervised learning to tabular domain}.
\newblock \bibinfo{journal}{Advances in Neural Information Processing Systems} \bibinfo{volume}{33}, \bibinfo{pages}{11033--11043}.
%Type = Article
\bibitem[{Zhang et~al.(2023)Zhang, Bengio and Singer}]{zhang2023layers}
\bibinfo{author}{Zhang, C.}, \bibinfo{author}{Bengio, S.}, \bibinfo{author}{Singer, Y.}, \bibinfo{year}{2023}.
\newblock \bibinfo{title}{Are all layers created equal?}
\newblock \bibinfo{journal}{Journal of Machine Learning Research} \bibinfo{volume}{24}, \bibinfo{pages}{1--52}.
%Type = Article
\bibitem[{Zhang et~al.(2021)Zhang, Bao and Ma}]{zhang2021self}
\bibinfo{author}{Zhang, L.}, \bibinfo{author}{Bao, C.}, \bibinfo{author}{Ma, K.}, \bibinfo{year}{2021}.
\newblock \bibinfo{title}{Self-distillation: Towards efficient and compact neural networks}.
\newblock \bibinfo{journal}{IEEE Transactions on Pattern Analysis and Machine Intelligence} \bibinfo{volume}{44}, \bibinfo{pages}{4388--4403}.
%Type = Inproceedings
\bibitem[{Zhang et~al.(2019)Zhang, Song, Gao, Chen, Bao and Ma}]{zhang2019be}
\bibinfo{author}{Zhang, L.}, \bibinfo{author}{Song, J.}, \bibinfo{author}{Gao, A.}, \bibinfo{author}{Chen, J.}, \bibinfo{author}{Bao, C.}, \bibinfo{author}{Ma, K.}, \bibinfo{year}{2019}.
\newblock \bibinfo{title}{Be your own teacher: Improve the performance of convolutional neural networks via self distillation}, in: \bibinfo{booktitle}{Proceedings of the IEEE/CVF International Conference on Computer Vision}, pp. \bibinfo{pages}{3713--3722}.
%Type = Inproceedings
\bibitem[{Zhang and Bifet(2020)}]{zhang2020feat}
\bibinfo{author}{Zhang, W.}, \bibinfo{author}{Bifet, A.}, \bibinfo{year}{2020}.
\newblock \bibinfo{title}{Feat: A fairness-enhancing and concept-adapting decision tree classifier}, in: \bibinfo{booktitle}{Discovery Science: 23rd International Conference, DS 2020, Thessaloniki, Greece, October 19--21, 2020, Proceedings 23}, \bibinfo{organization}{Springer}. pp. \bibinfo{pages}{175--189}.
%Type = Inproceedings
\bibitem[{Zhang et~al.(2018)Zhang, Zhou, Lin and Sun}]{zhang2018shufflenet}
\bibinfo{author}{Zhang, X.}, \bibinfo{author}{Zhou, X.}, \bibinfo{author}{Lin, M.}, \bibinfo{author}{Sun, J.}, \bibinfo{year}{2018}.
\newblock \bibinfo{title}{Shufflenet: An extremely efficient convolutional neural network for mobile devices}, in: \bibinfo{booktitle}{Proceedings of the IEEE conference on computer vision and pattern recognition}, pp. \bibinfo{pages}{6848--6856}.
%Type = Article
\bibitem[{Zhu et~al.(2023)Zhu, Shi, Erickson, Li, Karypis and Shoaran}]{zhu2023xtab}
\bibinfo{author}{Zhu, B.}, \bibinfo{author}{Shi, X.}, \bibinfo{author}{Erickson, N.}, \bibinfo{author}{Li, M.}, \bibinfo{author}{Karypis, G.}, \bibinfo{author}{Shoaran, M.}, \bibinfo{year}{2023}.
\newblock \bibinfo{title}{Xtab: Cross-table pretraining for tabular transformers}.
\newblock \bibinfo{journal}{arXiv preprint arXiv:2305.06090} .
%Type = Inproceedings
\bibitem[{{\v{Z}}liobaite(2010)}]{vzliobaite2010change}
\bibinfo{author}{{\v{Z}}liobaite, I.}, \bibinfo{year}{2010}.
\newblock \bibinfo{title}{Change with delayed labeling: When is it detectable?}, in: \bibinfo{booktitle}{2010 IEEE international conference on data mining workshops}, \bibinfo{organization}{IEEE}. pp. \bibinfo{pages}{843--850}.

\end{thebibliography}


\end{document}